\documentclass{article}

\usepackage{microtype}
\usepackage{graphicx}
\usepackage{subcaption}
\usepackage{booktabs} 
\usepackage{csquotes}
\usepackage{url}
\usepackage{comment}
\usepackage{paralist}
\usepackage{tabularx,array,makecell}
\usepackage{enumitem}
\usepackage{mathtools} 
\usepackage{amsthm}    
\usepackage{enumitem}  
\usepackage{booktabs}  
\usepackage[most]{tcolorbox}
\usepackage[ruled,vlined,linesnumbered]{algorithm2e}

\newtcolorbox{promptbox}[1][]{
    colback=black!1!white, 
    colframe=black!2!white, 
    fonttitle=\bfseries,
    coltitle=black,
    title=#1,
    breakable, 
    left=6mm,
    enhanced,
    attach boxed title to top left={yshift=-2mm, xshift=4mm},
    boxed title style={
        colback=black!5!white,
    }
}

\usepackage{hyperref}

\newcommand{\mypara}[1]{\textbf{#1.}}

\usepackage[accepted]{icml2026}

\usepackage{amsmath}
\usepackage{amssymb}
\usepackage{mathtools}
\usepackage{amsthm}

\usepackage[capitalize,noabbrev]{cleveref}

\theoremstyle{plain}

\theoremstyle{definition}

\theoremstyle{remark}

\usepackage[textsize=tiny]{todonotes}

\icmltitlerunning{Sparse Autoencoders are Topic Models}

\begin{document}

\twocolumn[
  \icmltitle{Sparse Autoencoders are Topic Models}



  \icmlsetsymbol{equal}{*}

  \begin{icmlauthorlist}
    \icmlauthor{Leander Girrbach}{aff}
    \icmlauthor{Zeynep Akata}{aff}
  \end{icmlauthorlist}

  \icmlaffiliation{aff}{Technical University of Munich (TUM), Munich Center for Machine Learning (MCML), Helmholtz Munich}
  \icmlcorrespondingauthor{Leander Girrbach}{leander.girrbach@tum.de}

  \icmlkeywords{sparse autoencoder, SAE, topic model}

  \vskip 0.3in
]



\printAffiliationsAndNotice{}  

\begin{abstract}
Sparse autoencoders (SAEs) are used to analyze embeddings, but their role and practical value are debated. We propose a new perspective on SAEs by demonstrating that they can be naturally understood as topic models. We propose a continuous topic model (CTM) inspired by Latent Dirichlet Allocation (LDA) for embedding spaces and derive the SAE objective as a maximum a posteriori estimator under this model. This view implies SAE features are thematic components rather than steerable directions. To confirm our theoretical findings, we introduce SAE-TM, a topic modeling framework that: (1) trains an SAE to learn reusable topic atoms, (2) interprets them as word distributions on downstream data, and (3) merges them into any number of topics without retraining. SAE-TM yields more coherent topics than strong baselines on text and image datasets while maintaining diversity. Finally, we analyze thematic structure in image datasets and trace topic changes over time in Japanese woodblock prints. Our work positions SAEs as effective tools for large-scale thematic analysis across modalities. Code is available at \href{https://github.com/ExplainableML/SAE-TM}{ExplainableML/SAE-TM}.
\end{abstract}

\section{Introduction}
\label{sec:intro}

Sparse autoencoders (SAEs) are an important tool for understanding embedding spaces, particularly the internal activations of foundation models \citep{bricken2023towards,kim2025interpreting}. However, practical, high-impact applications have remained limited, and SAEs have been criticized because of failures in steering \citep{wu2025axbench,peng2025use} and for being inferior to linear probes \citep{smith2025negative}. This raises important questions: (i) how should we understand SAEs, and (ii) how can we best use their strengths?

In this paper, we argue that SAEs are naturally understood as topic models, i.e.\ models that represent each datapoint as a mixture of prominent themes found in the entire dataset. Concretely, we propose a continuous topic model (CTM) inspired by Latent Dirichlet Allocation \citep[LDA]{blei2003latent} for embedding spaces and derive the SAE objective as a MAP estimator under this model. This implies that SAE features should be seen as thematic clusters whose activations combine to explain an embedding, rather than a monosemantic, steerable mechanism. Consequently, SAEs may indeed be less suited for mechanistic control at the level of single features. Instead, they work well for discovering and organizing unknown themes in data. We operationalize this view by constructing topic models directly from SAEs: we pretrain once to learn reusable topic atoms, interpret them as word distributions on downstream datasets, and merge them into any desired number of topics. We then evaluate the resulting models against strong baselines in both text and image settings.

\begin{figure}[t]
\centering
\includegraphics[width=\linewidth]{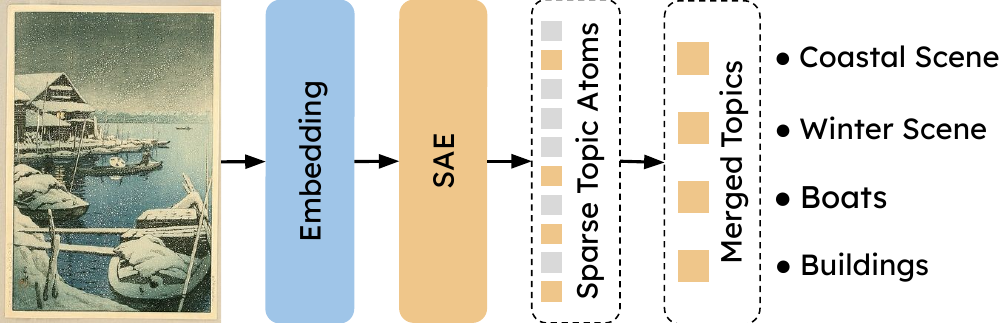}
\caption{We show theoretically and practically that Sparse Autoencoders are strong topic models in embedding space. An embedding model converts documents, such as images or text, into embeddings. An SAE then encodes these to produce sparse topic atoms. These atoms are merged into coarse-grained topics that are relevant to the entire dataset. Finally, we show that SAE Topic Models can be successfully used to analyze large-scale data.
}
\label{fig:teaser}
\vspace{-15pt}
\end{figure}

\begin{figure*}[t]
\centering
\includegraphics[width=\textwidth]{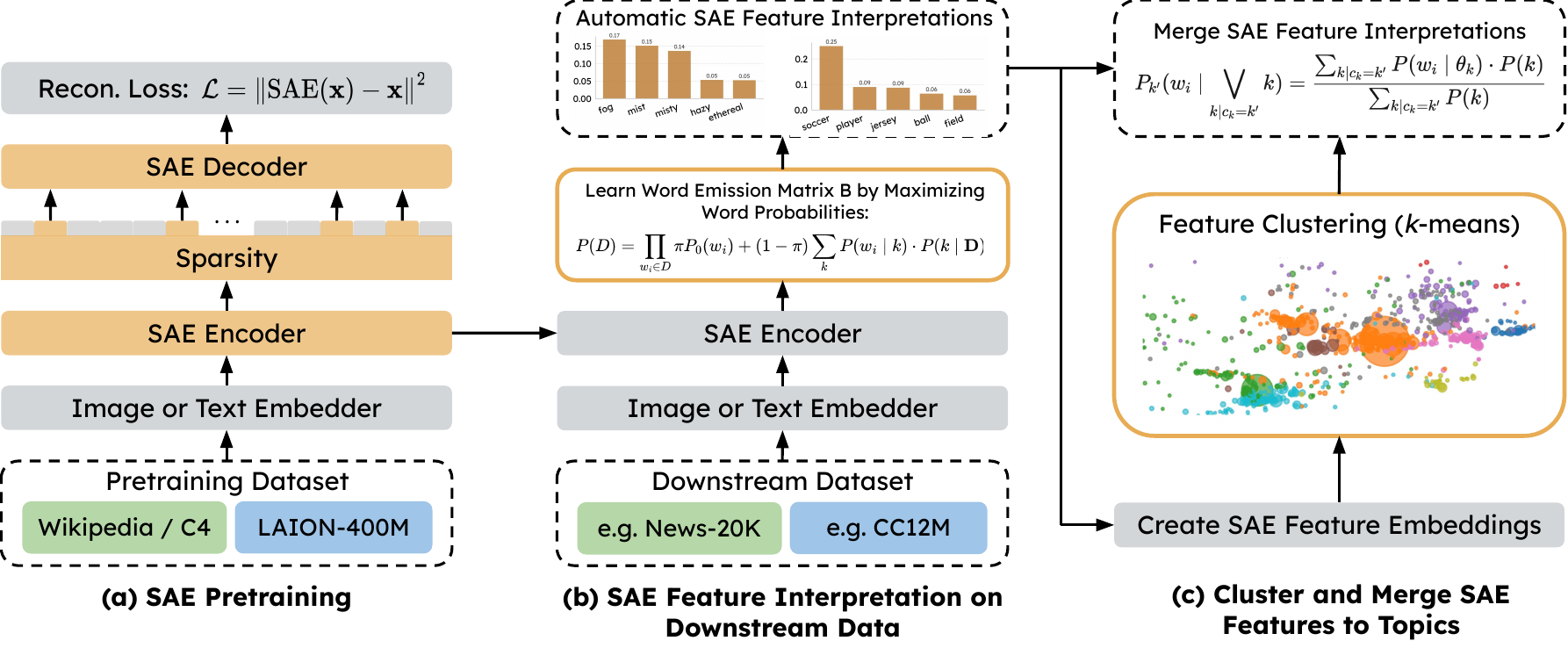}
\caption{Overview of our SAE topic model (SAE-TM): (a) pretrain foundational SAEs on large text or vision datasets to learn transferable atomic directions; (b) interpret relevant SAE features on downstream datasets by associating each feature with a distribution over words; (c) cluster SAE feature embeddings derived from their top associated words via $k$-means and merge clustered features into topics, aggregating their word distributions. Colors indicate modality (green = text, blue = vision) and trainable (orange) vs.\ frozen (grey) components.}
\label{fig:modelfigure}
\end{figure*}

In this way, SAE topic models also enable large-scale thematic analysis of image datasets. Although such datasets are central to computer vision research and applications, their broader thematic content is underexplored. By representing images as mixtures of learned topic atoms, SAEs enable efficient inspection of recurring visual themes within and across datasets. This offers a new perspective on dataset differences and similarities. We apply our approach to four widely used image datasets and find systematic and interpretable contrasts in their thematic structure.

In summary, our contributions are as follows: (1) we formalize a connection between topic models and SAEs by introducing a continuous topic model inspired by LDA and deriving the SAE objective as a MAP estimator under this model; (2) we show how to use SAEs as foundational topic models and find that they compare favorably to strong baselines on standard coherence and diversity metrics; (3) we apply our SAE-based topic model to analyze differences in the composition of four popular large-scale image datasets, finding clear and interpretable differences in their thematic structure, such as object-centric vs.\ human-centric emphases; and (4) we demonstrate how our SAE-TMs can detect changes in themes in Japanese woodblock prints across periods.
An overview is in \cref{fig:modelfigure}.

\section{Related Work}
\label{sec:related}

\begin{table*}[t]
  \centering
  \setlength{\tabcolsep}{6pt}
  \renewcommand{\arraystretch}{1.15}
  \small
  \begin{tabularx}{\linewidth}{
    >{\raggedright\arraybackslash}p{.48\linewidth}
    >{\raggedright\arraybackslash}p{.48\linewidth}
  }
    \toprule
    \textbf{(a) Classical LDA (discrete)} &
    \textbf{(b) Continuous Topic Model (embeddings)} \\
    \midrule

    \textbf{Hyperparams}:\; number of topics $K$;\; Dirichlet $\alpha\!\in\!\mathbb{R}^K_{>0}$;\; word dists $\beta\!\in\![0,1]^{K\times V}$ with rows on the simplex;\; doc-length rate $\xi$. &
    \textbf{Hyperparams}:\; $K$;\; Dirichlet $\alpha\!\in\!\mathbb{R}^K_{>0}$;\; directions $\mu_{1:K}\!\in\!\mathbb{R}^d$;\; covariances $\Sigma_{1:K}\!\succeq\!0$;\; strength dists $\operatorname{Ga}_{1:K}$ on $\mathbb{R}_{\ge0}$;\; Poisson rate $\rho_d$;\; noise $\sigma^2$. \\

    \addlinespace[2pt]
    \textbf{1. Topic mix}:\; $\theta \sim \mathrm{Dir}(\alpha)$. &
    \textbf{1. Topic mix}:\; $\theta \sim \mathrm{Dir}(\alpha)$. \\

    \addlinespace[2pt]
    \textbf{2. Length}:\; $N \sim \mathrm{Pois}(\xi)$. &
    \textbf{2. \# contributions}:\; $N \sim \mathrm{Pois}(\rho_d)$. \\

    \addlinespace[2pt]
    \textbf{3. For $n=1{:}N$}: &
    \textbf{3. For $n=1{:}N$}: \\
    \quad (a) topic $z_n \sim \mathrm{Cat}(\theta)$; &
    \quad (a) topic $z_n \sim \mathrm{Cat}(\theta)$; \\
    \quad (b) word $w_n \sim \mathrm{Cat}(\beta_{z_n})$. &
    \quad (b) direction $w_n \sim \mathcal{N}(\mu_{z_n},\Sigma_{z_n})$; \\
    &
    \quad (c) strength $\lambda_n \sim \operatorname{Ga}_{z_n}$; \\
    &
    \quad (d) contribution $c_n = \lambda_n\, w_n$. \\

    \addlinespace[2pt]
    \textbf{4. Document (obs.)}:\; bag-of-words counts $X=\sum_{n=1}^N e_{w_n}$. &
    \textbf{4. Embedding (obs.)}:\; $D=\sum_{n=1}^N c_n + \varepsilon$, with $\varepsilon\sim \mathcal{N}(0,\sigma^2 I)$. \\

    \addlinespace[2pt]
    \textbf{Mean (given $\theta$)}:\; $\mathbb{E}[X\,|\,\theta]=\beta^{\!\top}\theta$. &
    \textbf{Mean (given $\theta$)}:\; $\mathbb{E}[D\,|\,\theta]=W\theta$, where $W=[\rho_d m_1\mu_1,\dots,\rho_d m_K\mu_K]$, $m_k=\mathbb{E}[\lambda\,|\,z{=}k]$. \\
    \bottomrule
  \end{tabularx}
  \caption{Side-by-side comparison of the generative processes for classical LDA (discrete) and the proposed continuous topic model (CTM) for embedding spaces inspired by LDA. Steps are aligned to highlight both the shared structure and the differences.}
  \label{tab:gen-comparison}
\end{table*}

\mypara{Topic Modeling}
Most topic models infer topics as latent variables by maximizing the likelihood of the data. Early examples are LDA \citep{blei2003latent} and pLSI \citep{hofmann1999probabilistic}. These have inspired many extensions, including leveraging document information \citep{blei2006correlated}, dynamic topics \citep{blei2006dynamic}, and minibatch training \citep{hoffman2010online}. In Neural Topic Models (NTMs), \citet{srivastava2017autoencoding} combine LDA and VAEs \citep{kingma2013auto}, replacing the Dirichlet prior of LDA with a logistic normal distribution, which has become a standard approach of NTMs \citep{miao2016neural,miao2017discovering,card2018neural,burkhardt2019decoupling,dieng2020topic}. Other works improve topic quality, e.g., using a Wasserstein autoencoder \citep{nan2019topic}, a Weibull VAE to mitigate Gaussian latent problems \citep{zhang2018whai}, extend to multimodal data \citep{abaskohi2025cemtm}, or increase sparsity \citep{lin2019sparsemax,martins2016softmax}. Beyond VAEs, optimal transport infers word-topic mappings via transport plans between embeddings \citep{zhao2021neural,dan2022representing,wu2023effective,wu2024fastopic}. Other methods use contrastive learning \citep{nguyen2021contrastive} or prompt LLMs \citep{pham2024topicgpt}. Clustering document embeddings is also popular, where each cluster forms a topic \citep{aharoni2020unsupervised,sia2020tired,angelov2020top2vec,thompson2020topic,grootendorst2022bertopic,zhang2022neural}.
The main weaknesses of current NTMs are suffering from posterior collapse \citep{bowman2016generating,subramanian2018towards}, having an inflexible number of topics, and being heavily tailored towards text analysis. Our SAE-TM directly addresses those weaknesses.

\mypara{Sparse Autoencoders}
Sparse coding learns an overcomplete basis, using only a fraction of basis vectors per data point. Sparse Autoencoders (SAEs) have been shown to learn expressive, disentangled features \citep{olshausen1996emergence,ranzato2006efficient,ranzato2007sparse,makhzani2014k}. The ability of SAEs to learn sparse, independent directions enables LLM activation interpretation \citep{cunningham2023sparse,bricken2023towards} by extracting \enquote{monosemantic} directions \citep{elhage2022toy,klindt2025superposition}. We leverage this property to extract topic atoms from superimposed topics in document embeddings. Recent SAE improvements target expressivity \citep{rajamanoharan2024jumping,nabeshima2024matryoshka}, sparsity \citep{bussmann2024batchtopk,gao2025scaling}, separability \citep{hindupur2025projecting,engels2025not,engels2025decomposing}, and feature hierarchies \citep{fel2025archetypal,muchane2025incorporating}.
Recent work \citep{jiang2025interpretable,choi2025conceptscope} also proposes using SAEs for dataset inspection, but does not make a link to topic modeling. \citet{zheng2025model} adapt existing topic modeling methods to use SAE features as input tokens, while we theoretically establish that the SAE objective itself is a MAP estimator of a topic model.

\section{Comparing SAEs and Topic Models}
\label{sec:method}

\subsection{Background on Latent Dirichlet Allocation}
\label{sec:method:lda}

Latent Dirichlet Allocation (LDA) is a generative probabilistic model for collections of discrete data such as text corpora \citep{blei2003latent}. Each document is represented as a mixture of latent topics, where each topic is a distribution over words. Formally, for a document $w=(w_1,\dots,w_N)$ in a corpus with vocabulary size $V$ and $K$ topics, the process is:
\begin{compactitem}
    \item Draw topic proportions $\theta \sim \mathrm{Dir}(\alpha)$.
    \item For each position $n=1,\dots,N$ (with $N \sim \mathrm{Pois}(\xi)$):
    \begin{compactitem}
        \item Sample topic $z_n \sim \mathrm{Cat}(\theta)$.
        \item Sample word $w_n \sim \mathrm{Cat}(\beta_{z_n})$, where $\beta$ is a $K\times V$ matrix of word distributions.
    \end{compactitem}
\end{compactitem}
The joint distribution is $p(\theta,z,w|\alpha,\beta) = p(\theta|\alpha)\prod_{n=1}^N p(z_n|\theta)p(w_n|z_n,\beta)$, and integrating over $\theta$ yields the marginal $p(w|\alpha,\beta)$. Thus LDA consists of corpus-level parameters $(\alpha,\beta)$, document-level topic proportions $(\theta_d)$, and word-level assignments $(z_{dn},w_{dn})$.

\subsection{A Continuous Topic Model Inspired by LDA}
\label{sec:method:continuouslda}

We formalize a Continuous Topic Model (CTM) analogous to LDA, but operating on document embeddings $D \in \mathbb{R}^d$ instead of words. This makes topic modeling possible for other domains, such as vision, in addition to language, where similar precedents exist that learn topic models on embedding spaces \citep{dieng2020topic,das2015gaussian}. Our core assumption is that each document embedding is a linear combination of topic-specific continuous directions, which is a direct instantiation of the linear representation hypothesis \citep{park2024the}, i.e., embeddings are linear mixtures of factors. Formally, we represent $D$ as
$D = \epsilon + \sum_{i=1}^{N} \lambda_ic_i$,
where $c_i$ are the directions, $\lambda_i$ are strength coefficients that determine the scale of each direction, and $\epsilon$ is the variation left unexplained by the directions.

This formulation adapts LDA's structure for embeddings: The $N$ contributions correspond to the number of words (or components in an abstract sense) in a document. For each contribution, we sample a single topic $z_n \sim \operatorname{Cat}(\theta)$ from a document-level categorical distribution, where $\theta \sim \operatorname{Dir}(\alpha)$. As in LDA, $\alpha$ is a corpus-level parameter, and each contribution is assumed to be generated from one topic.

Finally, instead of sampling a discrete word from a categorical distribution, we sample a scaled continuous topic vector $\lambda_n w_n$. We independently sample the strength $\lambda_n$ from a Gamma distribution $\operatorname{Ga}_{z_n}$ and the continuous direction $w_n$ from a Gaussian distribution $\mathcal{N}(\mu_{z_n}, \Sigma_{z_n})$. However, note that $\mu_k$ are topic directions in document-embedding space. They are not embeddings of vocabulary items, and the CTM does not include topic-word distributions. A comparison of LDA and our CTM is in \cref{tab:gen-comparison}.

The expected document is linear in $\theta$, $\mathbb{E}[D|\theta] = W\theta$ with columns $W_{\cdot k} \propto \mu_k$, directly paralleling $\mathbb{E}[w|\theta]=\beta^\top\theta$ in LDA. Thus, both models share the same linear mixture structure, but differ in observation models (multinomial vs. Gaussian) and the data domain (discrete vs. continuous).

The main difference between the CTM and LDA is the introduction of the strength $\lambda$. This parameter is necessary to cover the entire space $\mathbb{R}^d$ with a finite mixture of topic directions. It is also motivated by viewing distributions of $w_n \sim \mathcal{N}(\mu_k, \Sigma_k)$ as directions instead of centroids.

\subsection{Relating SAEs with \texorpdfstring{$L_1$}{L1} Penalty to the CTM}
\label{sec:method:l1sparsitysaederivation}

In this section, we show that the SAE objective with $L_1$ penalty \citep{bricken2023towards} arises as a MAP estimator under the CTM of \cref{sec:method:continuouslda} with the following assumptions:
\begin{compactitem}
    \item[\textbf{(A1)}] \textit{High-activity, small-contribution limit:} $\rho_d \to \infty$ and $\alpha_0 \to 0$, with $\rho_d \alpha_0 \to \kappa \in (0, \infty)$, so that many small-strength contributions accumulate to a finite total.
    \item[\textbf{(A2)}] \textit{Concentrated topic directions:} $\Sigma_k \to 0$, so each contribution $w_n$ aligns with the topic mean direction $\mu_{z_n}$.
    \item[\textbf{(A3)}] \textit{Independence across topics:} the per-topic aggregated strengths $S_k$ are mutually independent given $\theta$.
\end{compactitem}
Under (A1)--(A3), $L_1$-penalty SAEs arise as the MAP solution to the CTM in the high-activity, small-contribution limit, as we now derive. Fixed-sparsity SAEs such as TopK \citep{gao2025scaling} and BatchTopK \citep{bussmann2024batchtopk} admit an analogous derivation under (A2)--(A3) that replaces (A1) with a hard support constraint (see \cref{sec:supp:fixedsparsitysaederivation}).

Assume the strength of each continuous contribution follows a Gamma distribution with common rate $\beta>0$ and shape $\alpha_0 > 0$ across topics,
\begin{equation}
\lambda_{k,i} \sim \operatorname{Ga}(\alpha_0, \beta), \qquad k=1,\ldots,K.
\end{equation}
Let $N\to\infty$ while each individual contribution becomes small. Concretely, we consider the limit
\begin{equation}
\rho_d \to \infty,\quad \alpha_0 \to 0,\quad \text{with } \rho_d \alpha_0\;\to\; \kappa \in (0,\infty),
\label{eq:high-activity-scaling}
\end{equation}
so many small-strength contributions accumulate to a finite total. Writing $N_k \sim \mathrm{Pois}(\rho_d \theta_k)$ for the number of contributions to topic $k$, the aggregated strength for topic $k$,
\begin{equation}
S_k := \sum_{i=1}^{N_k} \lambda_{k,i},
\end{equation}
converges in distribution to
\begin{equation}
S_k \ \Rightarrow\ \operatorname{Ga}\left(\kappa \theta_k,\beta\right).
\end{equation}
Define the total strength and normalized topic weights
\begin{equation}
s := \sum_{k=1}^K S_k, \qquad \tilde{\theta}_k := \frac{S_k}{s}.
\end{equation}
By independence across topics, we obtain
\begin{equation}
s \ \Rightarrow\ \operatorname{Ga}(\kappa,\beta),
\qquad
\tilde{\theta} \ \Rightarrow\ \operatorname{Dirichlet}(\kappa \theta),
\end{equation}
and $(s,\tilde{\theta})\ \perp\!\!\!\perp\ \text{given }\theta$. When $\kappa$ is large (i.e., many small contributions), $\tilde{\theta}$ concentrates on the document-level mixture $\theta$, and we may work with the collapsed representation
\begin{equation}
\ s \sim \mathrm{Ga}(\kappa,\beta),\quad
\theta \sim \mathrm{Dir}(\alpha),\quad
D = sW\theta + \varepsilon,
\label{eq:l1-collapsed}
\end{equation}
with $\varepsilon\sim \mathcal{N}(0,\sigma^2 I)$ and $W=[\mu_1,\ldots,\mu_K]\in\mathbb{R}^{d\times K}$, corresponding to the limit $\Sigma_k\to 0$ so that contributions align with $\mu_k$.
Reparameterizing by
\begin{equation}
a_k := s,\;
\theta_k
\Leftrightarrow
s \ =\ \sum_{k=1}^K a_k = \lVert a\rVert_1,\;
\theta_k = \frac{a_k}{\sum_j a_j},
\label{eq:s-theta-a}
\end{equation}
the observation model in \cref{eq:l1-collapsed} becomes the standard SAE decoder $D \mid a \ \sim\ \mathcal{N}(W a, \sigma^2 I)$ with $a\ge 0$. With the Gamma prior $s \sim \mathrm{Ga}(\kappa,\beta)$ and Dirichlet prior $\theta\sim \mathrm{Dir}(\alpha)$, the negative log-posterior for a single embedding $D$ reads
\begin{multline}
\mathcal{L}(\theta,s)
\ =
\frac{1}{2\sigma^2}\lVert D - W a\rVert_2^2
+
\underbrace{\beta s + (1-\kappa)\log s}_{-\log p(s)} \\
+
\underbrace{\sum_{k=1}^K (1-\alpha_k)\log \theta_k}_{-\log p(\theta)}
+ \text{const}, \qquad a = s\theta.
\label{eq:map-l1}
\end{multline}
Choosing $\kappa=1$ and $\alpha_k=1$ yields an exponential prior on $s$ and uniform Dirichlet on $\theta$, giving the SAE objective with $L_1$ penalty \citep{bricken2023towards}
\begin{equation}
\ \mathcal{L}(a) = \frac{1}{2\sigma^2}\lVert D - W a\rVert_2^2 + \beta\lVert a\rVert_1 + \text{const}, \quad a\ge 0.
\label{eq:sae-l1}
\end{equation}
When $\kappa<1$, the additional $(1-\kappa)\log s$ term further encourages a smaller total mass $s$, while $\alpha_k<1$ promotes peaked usage within the active topics.

\begin{figure}[t]
\centering
\includegraphics[width=\linewidth]{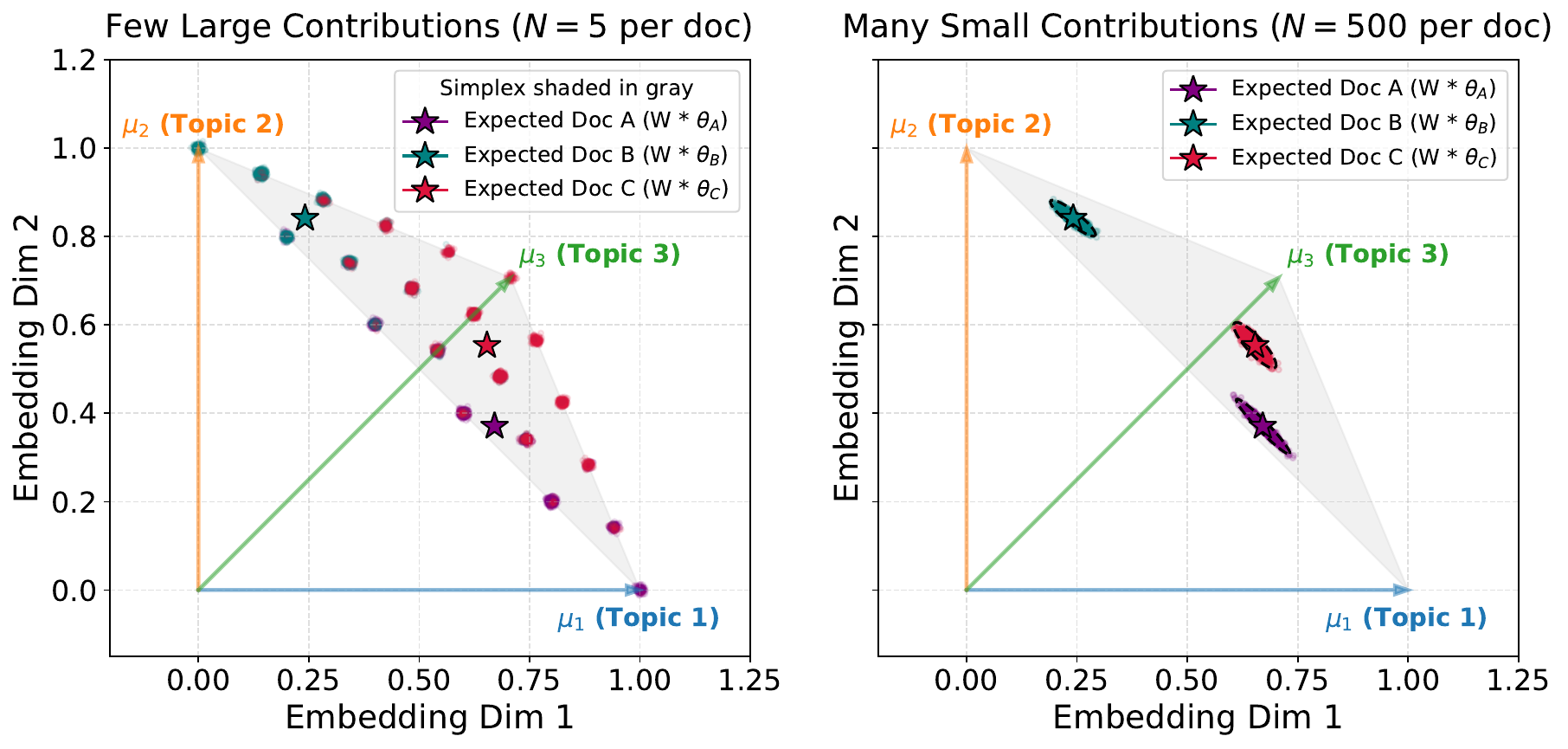}
\caption{Illustration of the high-activity, small-contribution limit (A1). We sample document embeddings from the CTM (\cref{sec:method:continuouslda}) for three documents $A$, $B$, $C$ with distinct topic mixtures $\theta_A, \theta_B, \theta_C$ over three topic directions $\mu_1, \mu_2, \mu_3$ (colored arrows; the topic simplex is shaded gray). Stars mark each document's expected embedding $W\theta$. \textbf{Left} ($N=5$, few large contributions): samples form a discrete, clumpy pattern that is poorly matched to the continuous $L_2$ reconstruction loss used by SAEs. \textbf{Right} ($N=500$, many small contributions): samples concentrate into Gaussian clouds around $W\theta$, recovering the SAE observation model of \cref{eq:l1-collapsed}.}
\label{fig:theoryvisualization}
\end{figure}

\mypara{Illustration}
To illustrate the theory between how SAEs constrain the high-activity, small-contribution limit, we show in \cref{fig:theoryvisualization} how embeddings for 3 documents are generated under different conditions. On the left, a document embedding is formed by only a few large topic contributions ($N=5$). In this case, possible embeddings form a rigid, clumpy grid. This discrete pattern doesn’t harmonize well with the continuous $L_2$ reconstruction loss used by SAEs. However, on the right, we move to the high-activity, small-contribution limit ($N=500$). There, the clumpy grid smoothes out into Gaussian clouds centered at each document's expected embedding ($W\theta$).

This explains how we can apply the standard SAE objective (i.e.\ the continuous $L_2$ loss) to a model based on discrete, word-like topic contributions (the CTM). At the same time, making each individual contribution small prevents the overall embedding size from increasing without bound, which mathematically yields the $L_1$ sparsity regularization.

\section{Applying SAEs as Topic Models}
\label{sec:saetopic}

\cref{sec:method} establishes that the SAE objective is a MAP estimator under the CTM: in the latent-variable sense, the SAE itself is the topic model, with each feature acting as a topic atom and its activation as the inferred per-document topic weight. To evaluate this topic model against standard baselines and to interpret its topics, however, two practical gaps must be bridged. First, SAEs typically have $\gg 1000$ features, while traditional topic models use a much smaller number of topics. Second, topics are commonly defined as a distribution over words \citep{srivastava2017autoencoding,wu2024fastopic}, an interpretation that SAEs do not directly support. We therefore introduce two post-hoc steps on top of the frozen SAE: (i) learning a word-emission matrix so that SAE features can be inspected as word distributions and evaluated with standard topic-model metrics, and (ii) merging features into a smaller set of topics to match the granularity used by baselines. Both steps operate on a frozen SAE and do not modify the generative model of \cref{sec:method}, so they are interpretation layers, not a new topic model.

\mypara{SAE topic interpretation}
We consider a trained SAE with a large number of latent features, $K \gg 1,000$. Our goal is to interpret these SAE features as topics, where each topic is a distribution over words, following conventions in topic modeling as noted above.
To interpret SAE features, we learn the word emission matrix $\mathbf{B} \in \mathbb{R}^{K \times V}$ by maximizing the bag-of-words likelihood of each datapoint given its SAE feature probabilities:
\begin{equation}\label{eq:saeinterpretation}
    P(D) = \prod_{w_i\in D}\pi P_0(w_i) + (1-\pi)\sum_{k} \underbrace{P(w_i\mid k)}_{=B_{k, i}}\cdot \underbrace{P(k\mid \mathbf{D})}_{=\theta_k},
\end{equation}
where $D$ is a textual representation of the document (for example, a detailed caption of an image), $\mathbf{D}$ is its embedding, and $P(\theta_k \mid \mathbf{D})$ is the normalized activation of the $k$-th SAE feature, as defined in \cref{eq:s-theta-a}. The background unigram prior $P_0(w)$ is an unconditional prior probability over words that accounts for words that are generally frequent but have no specific topic associations. This prevents the SAE feature interpretations from needing to model such common words. $\pi$ is set to $0.3$ in all experiments (see \cref{sec:supp:backgroundweightablation} for a systematic ablation). 
Additionally, we also improve SAE feature interpretations by weighting the contribution of each word to the document-level loss by its normalized inverse-document weight $\frac{\log\frac{N}{\text{df}(w_i)}}{\max_j \log\frac{N}{\text{df}(w_j)}}$, where $N$ is the total number of documents and $\text{df}(w_i)$ is the number of documents where word $w_i$ appears. We learn $\mathbf{B}$ by minimizing $-\log P(D)$ over the entire dataset.

\mypara{Topic merging}
As noted, SAEs typically have many features ($\gg 1,000$), while topic modeling requires fewer topics for better interpretability. We therefore treat SAE features as \textit{topic atoms} and construct broader topics from them. This approach has key advantages. The number of topics can be chosen flexibly using validation metrics without requiring retraining of the model, making this approach computationally effective. Additionally, we can directly assess the distinctness of the topics and use this information to decide which ones to merge.

Given $K$ SAE features and a target number of topics $K' < K$, we merge topics by creating embeddings for the SAE features and clustering them into $K'$ groups using $k$-means clustering. Topic embeddings are constructed as the weighted sum of word embeddings for the $V$ words in the vocabulary $\mathcal{V}$ used for feature interpretation. The topic embedding $\mathbf{T}_k$ is then defined as $\mathbf{T}_k = \sum_{w_i\in\mathcal{V}} B_{k, i}\cdot \mathbf{w}_i$, where $\mathbf{w}_i$ is the embedding of the word $w_i$. Additionally, we find it useful to denoise the topic embeddings by only considering the top-$p$ vocabulary \citep{holtzman2020curious}, i.e.\ taking the smallest set of words whose cumulative probability exceeds $p$ (we use $p=0.9$) and renormalizing so probabilities sum to 1.

For embeddings, we use widely available models like word2vec \citep{mikolov2013efficient} or GloVe \citep{pennington2014glove}. Alternatively, if word embeddings are not available, the SAE decoder weights for each feature also provide good topic embeddings \citep{rao2024discover}. After obtaining a cluster label $c_k$ for each topic, we merge the corresponding rows in $\mathbf{B}$ as follows:
\begin{equation}
    P_{k'}(w_i\mid \bigvee_{k\mid c_k = k'} k) = \frac{\sum_{k \mid c_k = k'} P(w_i\mid \theta_k)\cdot P(k)}{\sum_{k\mid c_k = k'} P(k)},
\end{equation}
where $P(k)$ is the average $\theta_k$ across all datapoints. \cref{sec:supp:topicmerging} contains an extended pseudocode and implementation details for the reader's convenience.

\mypara{Foundational SAE topic models}
Dynamically creating topics from topic atoms is also highly advantageous when working with limited data, similar to clustering pretrained embeddings. When building a topic model for a smaller dataset that cannot support training an expressive SAE from scratch, we can reuse the topic atoms from the pretrained SAE. We then use the statistics of the small dataset to select the relevant atoms and decide how to merge them.

\section{Evaluating SAE Topic Models}
\label{sec:experiments}

\begin{table*}[t]
\centering
\resizebox{\linewidth}{!}{
\begin{tabular}{lrrrrrrrrrrrrrrrrr}
\toprule
Num.\ Topics & \multicolumn{3}{c}{50} & \multicolumn{3}{c}{100} & \multicolumn{3}{c}{200} & \multicolumn{3}{c}{300} & \multicolumn{3}{c}{500} \\
\cmidrule(l){2-4} \cmidrule(lr){5-7} \cmidrule(lr){8-10} \cmidrule(lr){11-13} \cmidrule(r){14-16}
& $C_I \uparrow$ & $C_R \uparrow$ & $D \uparrow$ & $C_I \uparrow$ & $C_R \uparrow$ & $D \uparrow$ & $C_I \uparrow$ & $C_R \uparrow$ & $D \uparrow$ & $C_I \uparrow$ & $C_R \uparrow$ & $D \uparrow$ & $C_I \uparrow$ & $C_R \uparrow$ & $D \uparrow$ \\
\midrule
AVITM \citep{srivastava2017autoencoding} & 38.72 & 69.05 & 3.36 & 35.74 & \underline{69.02} & 3.36 & 34.59 & \underline{68.17} & 3.27 & \underline{33.38} & \underline{65.67} & 3.27 & 31.08 & \underline{59.61} & 3.13 \\
CombinedTM \citep{bianchi2021cross} & 40.90 & \underline{70.24} & 3.24 & \underline{38.49} & 67.37 & 3.22 & \underline{37.56} & 63.55 & 3.45 & 27.78 & 42.72 & 3.21 & \underline{31.79} & 50.77 & 3.24 \\
DecTM \citep{wu2021discovering} & 38.85 & 66.49 & 3.26 & 35.43 & 66.00 & 3.21 & 28.93 & 53.14 & 3.15 & 25.22 & 42.18 & 3.11 & 20.37 & 40.34 & 2.98 \\
DVAE \citep{burkhardt2019decoupling} & 21.24 & 22.45 & 3.00 & 17.36 & 11.16 & 3.36 & 16.87 & 2.29 & 3.21 & 16.64 & 4.69 & 3.20 & 16.50 & 2.87 & 3.17 \\
ETM \citep{dieng2020topic} & 21.68 & 28.36 & 3.08 & 20.26 & 23.28 & 3.18 & 20.02 & 19.05 & 3.25 & 19.77 & 17.91 & 3.29 & 19.23 & 15.62 & 3.34 \\
FASTopic \citep{wu2024fastopic} & 32.31 & 56.06 & 2.92 & 30.33 & 56.90 & 2.96 & 29.46 & 51.83 & 2.89 & 28.24 & 52.34 & 2.97 & 28.06 & 51.25 & 2.97 \\
NSTM \citep{zhao2021neural} & 21.73 & 39.62 & 3.07 & 22.88 & 38.95 & 3.04 & 22.61 & 41.69 & 2.99 & 22.59 & 42.49 & 2.95 & 23.43 & 48.58 & 2.76 \\
TSCTM \citep{wu2022mitigating} & \underline{44.61} & 69.75 & \textbf{3.87} & 35.81 & 58.53 & \textbf{3.79} & 29.51 & 40.00 & \textbf{3.76} & 26.17 & 27.40 & \textbf{3.70} & 21.68 & 17.67 & \textbf{3.70} \\
\midrule
SAE-TM (ours) & \textbf{54.31} & \textbf{77.25} & \underline{3.67} & \textbf{51.48} & \textbf{78.01} & \underline{3.64} & \textbf{46.63} & \textbf{75.71} & \underline{3.60} & \textbf{43.50} & \textbf{74.22} & \underline{3.59} & \textbf{40.49} & \textbf{71.22} & \underline{3.57} \\
\bottomrule
\end{tabular}
}
\caption{Results for topic modeling performance on five text datasets. Values show topic coherence ($C_I$ = intruder detection accuracy, $C_R$ = topic coherence rating) and diversity ($D$) scores. All scores are averaged over datasets. Best values are in bold, and second-best values are underlined. Different numbers of topics show trends when increasing topic granularity.}
\label{tab:results:textonly}
\vspace{-15pt}
\end{table*}

\mypara{Baselines}
We compare SAEs as topic models against representative state-of-the-art neural topic models: AVITM \citep{srivastava2017autoencoding}, CombinedTM \citep{bianchi2021cross,bianchi2021pre}, DecTM \citep{wu2021discovering}, DVAE \citep{burkhardt2019decoupling}, ETM \citep{dieng2020topic}, FASTopic \citep{wu2024fastopic}, NSTM \citep{zhao2021neural}, and TSCTM \citep{wu2022mitigating}. All implementations except DVAE are adapted from TopMost \citep{wu2024towards}. Among these models, only CombinedTM and FASTopic operate on embeddings (like SAEs). All other baselines require bag-of-words representations of documents as inputs. This highlights the necessity to develop embedding-based topic models for image applications. We do not include BERTopic \citep{grootendorst2022bertopic} because it targets a different definition of topic modeling: documents are clustered in embedding space, and each is assigned at most one topic. This constrasts to other baselines and also to SAE-TM, which assign multiple topics to each document, so results are not comparable.

\mypara{SAE Training}
All SAEs are trained and interpreted on individual datasets to match baselines (note that in \cref{fig:modelfigure} we adopt the \enquote{foundational SAE} perspective introduced above, but train SAEs on individual datasets here for fair comparison to baselines). We use the following hyperparameters: Expansion factor 4 (i.e.\ the dictionary size is 3072 for all models), $L_1$ penalty is $2$, batch size is $1000$, and we train SAEs for 50,000 steps with learning rate $0.001$. For interpretation, we use a vocabulary size 5000 (same for all baselines), batch size $1000$, and learning rate $0.01$. The number of epochs depends on the dataset (min.\ 50, max.\ 200). Training and interpretation are highly efficient, e.g.\ training an SAE on 50M embeddings (Twitter dataset) takes 10 minutes on a single GPU, interpretation 15 minutes.

\mypara{Evaluation Metrics}
Topic model evaluation is known to be challenging \citep{chang2009reading,lau2014machine,hoyle2021automated,harrando2021apples,rahimi2024contextualized}. The most important evaluation axes are \textit{Topic Coherence} and \textit{Topic Diversity} \citep{wu2024fastopic,wu2024survey}. Topic Coherence measures topic clarity and specificity, aiding interpretation. Topic Diversity measures inter-topic overlap and controls for models manipulating coherence by learning and repeating a few narrow topics.

We measure topic coherence using two metrics: Overall topic rating $C_R$ \citep{newman2010automatic} and intruder detection $C_I$ \citep{chang2009reading}. Overall rating scores the relatedness of a topic's top 20 words on a scale from 0 to 100. For intruder detection, we repeatedly sample 5 of the top-20 words from a topic and add an intruder word from a different topic. A judge must detect the intruder, and the score is the judge's accuracy. Following \citet{stammbach2023revisiting,rahimi2024contextualized}, we use LLMs as judges (\textsc{Phi-4}; \citet{abdin2024phi}), as prior work has established they align with human judges. Coherence scores are averages across topics.

We measure topic diversity by the avg.\ word mover distance \citep[WMD]{kusner2015word} between topics. For topics $k, k'$ represented by their top 20 words, WMD is defined as
\begin{equation}
    \operatorname{WMD}(k, k') = \min_{T\geq 0} \sum_{i, j} T_{i, j}C_{i, j},
\end{equation}
where $\quad C_{i,j} = \lVert \mathbf{w}_i - \mathbf{w}'_j\lVert_2$ is the distance of paired word embeddings and $T$ follows optimal transport constraints (uniform marginals). Intuitively, WMD calculates the \enquote{cost} to \enquote{move} all word embeddings from one topic to the other. WMD ranks models similarly to the ratio of unique words \citep{dieng2020topic}, but is robust to varying topic numbers (see \cref{sec:supp:topicdiversitymetric}).

Finally, we also demonstrate that SAE-TM does not only learn coherent and diverse topics, but topics are relevant to the respective documents. For this, we sample one active topic and one inactive topic per document and let an LLM decide if the topic is relevant to the document. See \cref{sec:supp:topicrelevance} for the results on this metric.

\subsection{Evaluation on Text-Only Datasets}
\label{sec:experiments:language}

\mypara{Datasets}
We evaluate topic models on five text-only datasets: News-20K \citep{mitchell1997machine}, IMDB movie reviews \citep{maas2011learning}, Yelp restaurant reviews \citep{zhang2015character}, DailyMail stories \citep{hermann2015teaching}, and a filtered tweet dataset \citep{cheng2010you}. These datasets cover diverse domains such as reviews (IMDB and Yelp), news (News-20K and DailyMail), and social media (Twitter). They also vary in size: News-20K has 18,846 documents, IMDB and Yelp each have 50,000, DailyMail has 219,507 articles, and the filtered Twitter dataset has 1,183,728 tweets. The original Twitter dataset contains over three million tweets, but we only keep those with at least seven non-stopword lemmas from the top 5,000 most common lemmas to ensure sufficient content for topic detection. We preprocess documents as follows: We lemmatize all words and filter stopwords. In each dataset, we determine the 5,000 most frequent lemmas and ignore all others. We use the \textsc{NLTK} library \citep{bird2006nltk} for tokenization, lemmatization, and stopword filtering. For document embeddings, we use \textsc{Granite-R2} \citep{awasthy2025granite}.

\mypara{Results}
\cref{tab:results:textonly} shows metrics averaged across text-only datasets, but for different numbers of topics. SAE-TM outperforms all baselines in topic coherence, achieving the highest scores for both intruder detection ($C_I$) and overall rating ($C_R$). Regarding topic diversity ($D$), SAE-TM consistently ranks second, trailing only TSCTM, which boosts diversity by upweighting semantically related, but low-frequency words. Additionally, the coherence of topics identified by TSCTM declines sharply as the number of topics increases, dropping from 69.75 ($C_R$) at 50 topics to 17.67 at 500 topics. SAE-TM, in contrast, maintains high and stable coherence scores even at 500 topics. The next-best performing baselines for coherence, typically AVITM and CombinedTM, still fall significantly short of the performance achieved by our SAE-TM.

\subsection{Evaluation on Image Datasets}
\label{sec:experiments:vision}

\begin{table*}[t]
\centering
\resizebox{\linewidth}{!}{
\begin{tabular}{lrrrrrrrrrrrrrrr}
\toprule
Num.\ Topics & \multicolumn{3}{c}{50} & \multicolumn{3}{c}{100} & \multicolumn{3}{c}{200} & \multicolumn{3}{c}{300} & \multicolumn{3}{c}{500} \\
\cmidrule(l){2-4} \cmidrule(lr){5-7} \cmidrule(lr){8-10} \cmidrule(lr){11-13} \cmidrule(r){14-16}
& $C_I \uparrow$ & $C_R \uparrow$ & $D \uparrow$ & $C_I \uparrow$ & $C_R \uparrow$ & $D \uparrow$ & $C_I \uparrow$ & $C_R \uparrow$ & $D \uparrow$ & $C_I \uparrow$ & $C_R \uparrow$ & $D \uparrow$ & $C_I \uparrow$ & $C_R \uparrow$ & $D \uparrow$ \\
\midrule
AVITM & 31.12 & 78.52 & 3.44 & 30.64 & 78.11 & 3.40 & 28.88 & \underline{76.78} & 3.36 & 28.40 & \underline{75.74} & 3.35 & 27.90 & \underline{74.06} & 3.36 \\
CombinedTM & \underline{42.30} & 79.39 & 3.82 & 35.11 & 59.65 & 3.82 & 23.16 & 30.80 & 3.80 & 21.62 & 28.74 & \textbf{3.83} & 20.29 & 26.56 & 3.80 \\
DecTM & 35.20 & 69.18 & \textbf{3.94} & 33.96 & 64.44 & \textbf{3.93} & 28.32 & 45.73 & \textbf{3.94} & 16.40 & 16.58 & 3.74 & 16.30 & 11.81 & \underline{3.81} \\
DVAE & 16.93 & 5.75 & 3.64 & 16.06 & 5.64 & 3.51 & 16.94 & 10.05 & 3.20 & 16.08 & 6.20 & 3.16 & 16.25 & 7.44 & 3.11 \\
ETM & 20.46 & 42.15 & 3.43 & 19.44 & 35.63 & 3.51 & 19.33 & 28.85 & 3.58 & 18.67 & 24.45 & 3.62 & 18.25 & 19.93 & 3.67 \\
FASTopic & 34.44 & 69.56 & 3.54 & 33.00 & 70.06 & 3.54 & 32.28 & 68.14 & 3.57 & 32.68 & 69.94 & 3.55 & \underline{31.05} & 67.27 & 3.53 \\
NSTM & 19.57 & 63.88 & 2.79 & 18.34 & 65.65 & 2.73 & 19.41 & 67.90 & 2.71 & 19.09 & 67.76 & 2.67 & 18.48 & 70.00 & 2.63 \\
TSCTM & 40.51 & \underline{80.40} & \underline{3.91} & \underline{38.46} & \underline{80.33} & \underline{3.84} & \underline{34.69} & 72.61 & \textbf{3.82} & \underline{30.15} & 57.39 & \underline{3.82} & 25.28 & 39.81 & \textbf{3.83} \\
\midrule
SAE-TM (ours) & \textbf{42.57} & \textbf{85.05} & 3.70 & \textbf{40.11} & \textbf{85.67} & 3.54 & \textbf{38.59} & \textbf{85.53} & 3.54 & \textbf{37.57} & \textbf{85.05} & 3.53 & \textbf{36.54} & \textbf{84.43} & 3.53 \\
\bottomrule
\end{tabular}
}
\caption{Performance on three image datasets. Values show topic coherence ($C_I$ = intruder detection accuracy, $C_R$ = topic coherence rating) and diversity ($D$) scores. All scores are averaged over datasets. Best values are in bold, and second-best values are underlined.}
\label{tab:results:imagedataset}
\end{table*}

\mypara{Datasets}
We evaluate topic models on three image datasets: CIFAR100 \citep{krizhevsky2009learning} (50,000 images), Food101 \citep{bossard14} (75,750 images), and SUN397 \citep{xiao2010sun} (108,618 images). To embed images, we use \textsc{ViT-B-16-SigLIP} from OpenCLIP \citep{ilharco2021openclip}.
We also create detailed, long-form captions by \textsc{InternVL3.5-14B} \citep{wang2025internvl3} for all images, which are necessary for learning topic models. For baselines that only process text data, we make a best-effort comparison by training them on captions. However, SAEs, FASTopic, and CombinedTM directly use image embeddings. Caption preprocessing for learning word emission probabilities follows \cref{sec:experiments:language}. SAEs furthermore support alternative interpretation methods for images, such as top-$k$ most activating images per feature with subsequent description or keyword extraction, however we use only captions here to allow a fair comparison to baselines.

\mypara{Results}
\cref{tab:results:imagedataset} shows metrics averaged across three image datasets. Similar to text data, SAE-TM identifies significantly more coherent topics than other methods, which confirms its strong performance in analyzing image data as well. However, its topic diversity is weaker than some baselines but comparable to its performance on text datasets. This means other methods can increase the diversity of their topics on image datasets, while SAE-TM remains stable. We attribute this to modality effects, as image embeddings focus on a few foreground objects \citep{li2025unbiased,xiao2025flair,kar2024brave}, and SAE features sometimes represent concepts that cannot be explained well in language, subsequently binding high-frequency words in interpretation. However, this performance gap warrants further improvements to SAE feature interpretation to focus on topic-relevant words. As an orthogonal contribution, we observe that other methods that operate on image embeddings (CombinedTM and FASTopic) also yield good topics. CombinedTM performs well for a low number of topics, whereas FASTopic remains stable across different numbers of topics. This confirms the potential for learning TMs using image embeddings alone.

\section{Inspecting Image Datasets with SAE TMs}
\label{sec:applications}

\subsection{Topic Distribution of Popular Datasets}
\label{sec:applications:imagedatasets}

We analyze the most prevalent topics and topic differences in popular image datasets, i.e.\ ImageNet \citep{deng2009imagenet}, CC3M \citep{sharma2018conceptual}, CC12M \citep{changpinyo2021conceptual}, and YFCC-15M \citep{thomee2016yfcc100m}. These datasets are widely used, but their content remains underexplored beyond concept frequency statistics \citep{garcia2023uncurated,udandarao2024no,wiedemer2024pretraining}. Hence, we utilize  SAE topic models to examine diverging themes in these datasets and understand their differences.

\mypara{Dataset Preprocessing}
The combined datasets contain $>$ 30 million images. To interpret the SAE topic model's features, we pair each image with a caption by \textsc{InternVL3.5-14B}. We follow the caption processing described in \cref{sec:experiments:language} and embed images with \textsc{ViT-B-16-SigLIP}, and use the foundational SAE described in \cref{sec:supp:saepretraining}.

\mypara{Deriving topics}
We learn word emission probabilities for SAE features on the combined 30 million images, as described in \cref{sec:saetopic}. We then cluster the selected SAE features in 100 topics (topic merging, see \cref{sec:saetopic}). 
After merging, we classify topics into two categories. Abstract topics are concerned with general image properties, such as mood, perspective, geometry, or layout. Concrete topics are concerned with objects visible in the images. In our analysis, we focus on concrete topics. For classification, we use an LLM (\textsc{Phi-4}; \citet{abdin2024phi}). The same LLM also summarizes the top 20 words in each topic by emission probability. Additionally, we remove topics that activate for more than 30\% of images (macro-average).

\begin{figure*}[t]
\centering
\includegraphics[trim=0pt 8pt 0pt 0pt, clip, width=\linewidth]{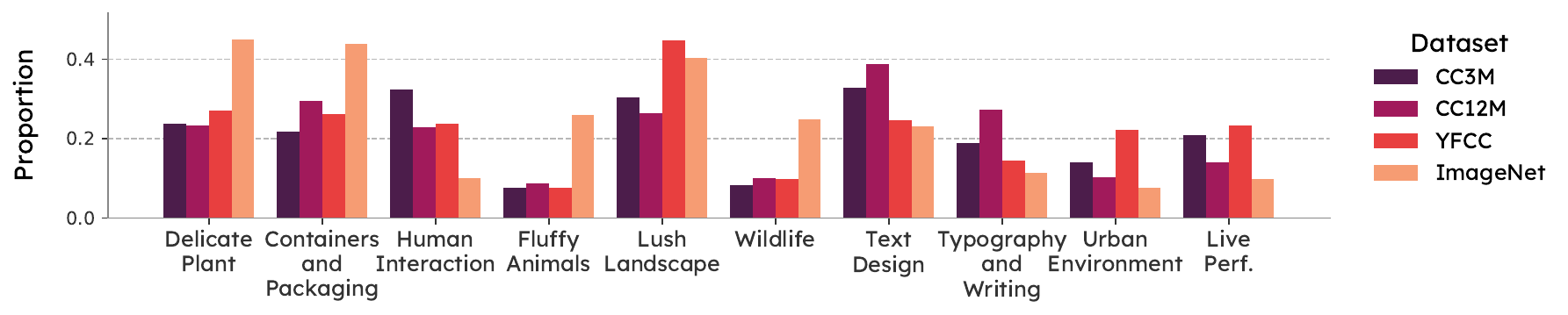}
\caption{Statistics of top 10 topics with the highest variance across four popular image datasets. Values indicate the proportion of images in each dataset where the topic is active (even weakly). Differences between datasets reveal interesting trends, such as a comparatively higher frequency of images of animals and plants in ImageNet compared to web-sourced datasets.}
\label{fig:imagedatasetcomposition}
\end{figure*}

\mypara{Results}
In \cref{fig:imagedatasetcomposition}, we show the top 10 topics sorted by the variance of their activity ratio across datasets. ImageNet has significantly more plants (\enquote{Delicate Plants}) and animals (\enquote{Fluffy Animals}, \enquote{Wildlife}) than other datasets. ImageNet also has more technical tools (\enquote{Containers and Packaging}). In contrast, ImageNet has significantly fewer images showing humans (\enquote{Human Interaction}). These differences reflect the dataset construction. The class-balanced approach of ImageNet emphasizes animals and common objects, which are less frequent in the other three web-based datasets. However, differences also exist among them: CC3M and CC12M contain more text and typographic elements, a trend that is particularly pronounced for CC12M. YFCC features many urban scenes (\enquote{Urban Environment}) and, with CC3M, many musical performances (\enquote{Live Performance}). Like ImageNet, YFCC also features many natural landscapes (\enquote{Lush Landscape}). These results show that dataset differences arise from construction, but even with similar methodology (CC3M, CC12M), differences emerge from varying image sources. Overall, we find that Sparse Autoencoders are an effective and efficient tool for understanding dataset composition, which directly motivates applications to trace downstream behavior, dataset rebalancing, and data selection. Importantly, SAE topic models avoid the need for expensive attribute labeling via MLLMs or specialized models \citep{zhang2024recognize,huang2023open}. Our analysis can be easily expanded by considering more or finer-grained topics, down to atomic SAE features.

\begin{figure*}[t]
\centering
\includegraphics[trim=0pt 10pt 0pt 0pt, clip, width=\linewidth]{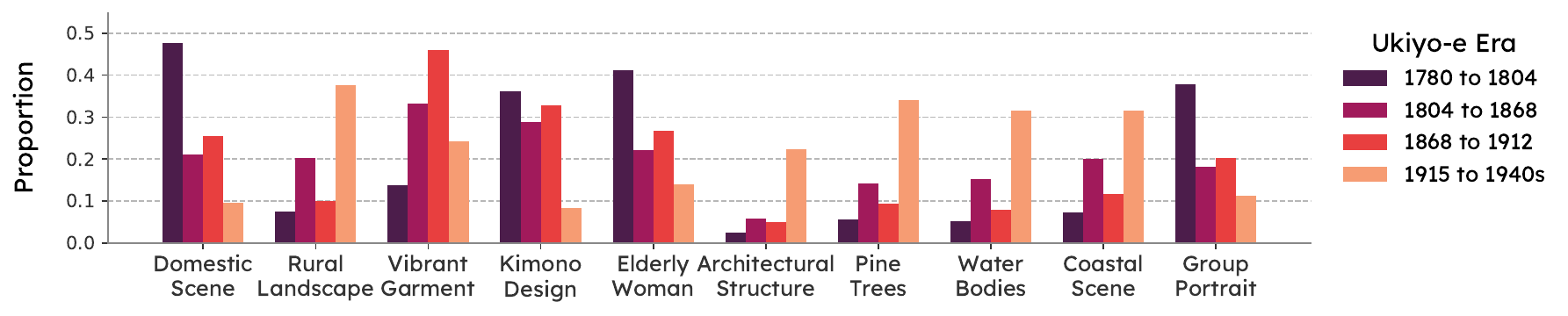}
\caption{Statistics of top 10 topics with the highest variance in Japanese woodblock prints from different artistic periods. Changes in topic distribution reflect changing cultural environment (e.g., clothing) and popular themes (e.g., domestic scenes vs.\ nature).}
\label{fig:ukiyoecomposition}
\end{figure*}

\subsection{Analyzing Topic Distribution in Artworks}
\label{sec:applications:art}

We analyze how prevalent topics evolve in Japanese woodblock prints (177,897 images) provided by \citet{khan2021stylistic}. This demonstrates applications of topic modeling in the humanities, specifically how deep learning research can enhance our understanding of cultural assets. We create captions for all images using InternVL3.5-14B and apply our pretrained text SAE to \textsc{Granite-R2} embeddings of the captions, as we find that complex themes in art are poorly represented by SigLIP embeddings.

\mypara{Results}
As in \cref{sec:applications:imagedatasets}, we show the 10 topics with the highest variance across categories. In \cref{fig:ukiyoecomposition}, we present the resulting topic proportions for four of the seven eras (images are categorized into seven eras spanning from the 1740s to the present) to illustrate trends (results for eras earlier than 1800 and later than 1950 are not shown to enhance clarity, but follow the observed trends).

Topics can be subdivided into four groups: Domestic scenes (\enquote{Domestic Scene}, \enquote{Elderly Woman}, and \enquote{Group Portrait}), Nature (\enquote{Rural Landscape}, \enquote{Pine Trees}, \enquote{Water Bodies}, and \enquote{Coastal Scene}), Building (\enquote{Architectural Structure}), and Fashion/Accessories (\enquote{Vibrant Garment} and \enquote{Kimono Design}). Depictions of domestic scenes are particularly prevalent in the \enquote{Golden Age of Ukiyo-e (1780 to 1804)}, and continually declined in popularity afterwards. Regarding fashion design, a clear divide is evident between the 20th century and the Edo and Meiji periods. Traditional attire is depicted much less frequently in the 20th century than before, reflecting changing customs and increasing Western influence.
Finally, depictions of natural scenes and architecture experienced a significant increase in popularity in 20th-century woodblock prints. However, they were also relatively more popular at the end of the Edo period (1804-1868) compared to earlier periods and remained popular during the Meiji era. This reflects a shift from woodblock prints that focused on human relations and social scenes to those that featured nature and landscapes.

This analysis already reveals interesting trends that can be further expanded by considering more fine-grained topics, which is easily achieved given the modular design of SAE-derived topics. In summary, topic modeling is a useful tool for understanding complex image datasets, such as art, and our proposed SAE topic model is a well-suited method.

\section{Conclusion}
\label{sec:conclusion}

We have shown that SAEs can be understood as topic models by introducing a continuous topic model inspired by LDA and deriving the SAE objective as the corresponding MAP estimator under specific conditions. Under this view, SAE features function as thematic components that combine to explain embeddings, rather than as monosemantic or steerable mechanisms. Building on this, we proposed SAE-TMs, a framework in which SAEs are pretrained once to learn reusable topic atoms and later interpreted and merged for downstream datasets.
Our experiments show that SAE-TMs yield coherent and diverse topics across five text and three image datasets, outperforming strong Neural Topic Modeling baselines. We further use SAE-TMs to compare thematic structure across image datasets and to analyze changes in themes in woodblock prints.

Concretely, SAE-TMs make five contributions to topic modeling. First, ours is the first derivation that casts the SAE objective as MAP inference under a topic model, providing a principled grounding for SAEs as topic-discovery tools. Second, the topic count $K'$ can be varied freely after pretraining via merging, with no retraining required. Third, SAE-TMs operate entirely in latent embedding space. They reconstruct the embedding rather than a bag-of-words, making them easily applicable across modalities. Fourth, SAE-TMs decouple topic learning (the frozen SAE) from interpretation (post-hoc word-emission learning), giving practitioners flexibility in how topics are represented. Fifth, every merged topic is transparently composed of its fine-grained atoms, which supports interpretability at multiple scales.

However, some limitations remain. SAE feature interpretation can be improved, and activation strengths do not always align with topic importance. Document embeddings also encode non-thematic information, such as sentiment, document length, or stylistic features, which SAE-TMs may encode alongside thematic structure. However, this is a property shared by all embedding-based topic models, including CombinedTM and FASTopic, and is not specific to our approach.

Several extensions of the SAE-TM framework follow naturally from our analysis. Assumption (A3) of \cref{sec:method:l1sparsitysaederivation} restricts our derivation to SAE variants with independent features, but recent hierarchical SAEs, such as Matryoshka SAEs \citep{nabeshima2024matryoshka} that train nested codebooks at multiple scales, and Matching-Pursuit SAEs \citep{costa2025flat} that explicitly model hierarchical structure over directions, relax this assumption and suggest extensions toward hierarchical topic models. We see formalizing this connection and developing the corresponding inference procedure as a promising direction for future work.

Taken together, this work presents a unified theoretical framework, a practical modeling approach, and applications that underscore an effective role for SAEs in data analysis and representation research.

\section*{Impact Statement}
This paper presents foundational work on theoretically understanding Sparse Autoencoders with practical applications aimed at understanding the thematic structure of large-scale datasets. By formalizing the connection between SAEs and Topic Models, our research contributes to better insights into the properties of widely used interpretability tools and to the development of scalable methods to audit training data for potential biases and distributional differences. Furthermore, we demonstrate the utility of these methods in the digital humanities, supporting the analysis of cultural heritage materials.

However, there are ethical considerations regarding the reliability of automated interpretations. Our method relies on associating features with word distributions from existing textual descriptions or LLM-generated captions to assign meaning to topics. Biases or hallucinations present in these documents (e.g., stemming from captioning models but also biases of human annotators) could propagate into the resulting topic models, potentially leading to misrepresentations of data groups. Additionally, while our approach helps in discovering unknown themes, activation strengths do not always perfectly align with semantic importance. Therefore, users should exercise caution and perform qualitative validation when using these tools to draw conclusions about sensitive or sociological datasets.

\subsection*{Acknowledgements}
This work was partially funded by the ERC (853489 - DEXIM) and the Alfried Krupp von Bohlen und Halbach Foundation, for which we thank them for their generous support. The authors gratefully acknowledge the scientific support and resources of the AI service infrastructure \textit{LRZ AI Systems} provided by the Leibniz Supercomputing Centre (LRZ) of the Bavarian Academy of Sciences and Humanities (BAdW), funded by Bayerisches Staatsministerium für Wissenschaft und Kunst (StMWK).
We also acknowledge the use of the HPC cluster at Helmholtz Munich for the computational resources used in this study.

\bibliography{main}

@inproceedings{blei2003latent,
  title={Latent dirichlet allocation},
  author={Blei, David and Ng, Andrew and Jordan, Michael},
  booktitle={JMLR},
  year={2003}
}

@inproceedings{hofmann1999probabilistic,
  title={Probabilistic latent semantic indexing},
  author={Hofmann, Thomas},
  booktitle={SIGIR},
  year={1999}
}

@inproceedings{blei2006correlated,
  title={Correlated topic models},
  author={Blei, David and Lafferty, John},
  booktitle={NeurIPS},
  year={2005},
}

@inproceedings{hoffman2010online,
  title={Online learning for latent dirichlet allocation},
  author={Hoffman, Matthew and Bach, Francis and Blei, David},
  booktitle={NeurIPS},
  year={2010}
}

@inproceedings{blei2006dynamic,
  title={Dynamic topic models},
  author={Blei, David and Lafferty, John},
  booktitle={ICML},
  year={2006}
}

@inproceedings{kingma2013auto,
  title={Auto-encoding variational bayes},
  author={Kingma, Diederik P and Welling, Max},
  booktitle={ICLR},
  year={2014}
}

@inproceedings{srivastava2017autoencoding,
  title={Autoencoding Variational Inference For Topic Models},
  author={Srivastava, Akash and Sutton, Charles},
  booktitle={ICLR},
  year={2017}
}

@inproceedings{miao2017discovering,
  title={Discovering discrete latent topics with neural variational inference},
  author={Miao, Yishu and Grefenstette, Edward and Blunsom, Phil},
  booktitle={ICML},
  year={2017},
}

@inproceedings{miao2016neural,
  title={Neural variational inference for text processing},
  author={Miao, Yishu and Yu, Lei and Blunsom, Phil},
  booktitle={ICML},
  year={2016},
}

@inproceedings{nan2019topic,
  title={Topic Modeling with Wasserstein Autoencoders},
  author={Nan, Feng and Ding, Ran and Nallapati, Ramesh and Xiang, Bing},
  booktitle={ACL},
  year={2019}
}

@inproceedings{zhang2018whai,
  title={{WHAI}: Weibull Hybrid Autoencoding Inference for Deep Topic Modeling},
  author={Hao Zhang and Bo Chen and Dandan Guo and Mingyuan Zhou},
  booktitle={ICLR},
  year={2018},
}

@inproceedings{lin2019sparsemax,
  title={Sparsemax and relaxed wasserstein for topic sparsity},
  author={Lin, Tianyi and Hu, Zhiyue and Guo, Xin},
  booktitle={WDSM},
  year={2019}
}

@inproceedings{martins2016softmax,
  title={From softmax to sparsemax: A sparse model of attention and multi-label classification},
  author={Martins, Andre and Astudillo, Ramon},
  booktitle={ICML},
  year={2016},
}

@inproceedings{card2018neural,
  title={Neural Models for Documents with Metadata},
  author={Card, Dallas and Tan, Chenhao and Smith, Noah A},
  booktitle={ACL},
  year={2018}
}

@inproceedings{dieng2020topic,
  title={Topic modeling in embedding spaces},
  author={Dieng, Adji B and Ruiz, Francisco JR and Blei, David M},
  booktitle={TACL},
  year={2020},
}

@inproceedings{bianchi2021pre,
  title={Pre-training is a hot topic: contextualized document embeddings improve topic coherence},
  author={Bianchi, Federico and Terragni, Silvia and Hovy, Dirk and others},
  booktitle={ACL},
  year={2021},
}

@inproceedings{bianchi2021cross,
  title={Cross-lingual contextualized topic models with zero-shot learning},
  author={Bianchi, Federico and Terragni, Silvia and Hovy, Dirk and Nozza, Debora and Fersini, Elisabetta and others},
  booktitle={EACL},
  year={2021}
}

@inproceedings{dan2022representing,
  title={Representing Mixtures of Word Embeddings with Mixtures of Topic Embeddings},
  author={Guo, Dan and Zhao, He and Zheng, Huangjie and Tanwisuth, Korawat and Chen, Bo and Zhou, Mingyuan and others},
  booktitle={ICLR},
  year={2022}
}

@inproceedings{zhao2021neural,
  title={Neural topic model via optimal transport},
  author={Zhao, He and Phung, Dinh and Huynh, Viet and Le, Trung and Buntine, Wray},
  booktitle={ICLR},
  year={2021}
}

@inproceedings{wu2024fastopic,
  title={Fastopic: Pretrained transformer is a fast, adaptive, stable, and transferable topic model},
  author={Wu, Xiaobao and Nguyen, Thong and Zhang, Delvin and Wang, William Yang and Luu, Anh Tuan},
  booktitle={NeurIPS},
  year={2024}
}

@inproceedings{nguyen2021contrastive,
  title={Contrastive learning for neural topic model},
  author={Nguyen, Thong and Luu, Anh Tuan},
  booktitle={NeurIPS},
  year={2021}
}

@inproceedings{pham2024topicgpt,
  title={TopicGPT: A Prompt-based Topic Modeling Framework},
  author={Pham, Chau and Hoyle, Alexander and Sun, Simeng and Resnik, Philip and Iyyer, Mohit},
  booktitle={NAACL},
  year={2024}
}

@inproceedings{aharoni2020unsupervised,
  title={Unsupervised Domain Clusters in Pretrained Language Models},
  author={Aharoni, Roee and Goldberg, Yoav},
  booktitle={ACL},
  year={2020}
}

@inproceedings{sia2020tired,
  title={Tired of Topic Models? Clusters of Pretrained Word Embeddings Make for Fast and Good Topics too!},
  author={Sia, Suzanna and Dalmia, Ayush and Mielke, Sabrina J},
  booktitle={EMNLP},
  year={2020}
}

@inproceedings{thompson2020topic,
  title={Topic modeling with contextualized word representation clusters},
  author={Thompson, Laure and Mimno, David},
  booktitle={arXiv},
  year={2020}
}

@inproceedings{grootendorst2022bertopic,
  title={BERTopic: Neural topic modeling with a class-based TF-IDF procedure},
  author={Grootendorst, Maarten},
  booktitle={arXiv},
  year={2022}
}

@inproceedings{zhang2022neural,
  title={Is Neural Topic Modelling Better than Clustering? An Empirical Study on Clustering with Contextual Embeddings for Topics},
  author={Zhang, Zihan and Fang, Meng and Chen, Ling and Namazi-Rad, Mohammad-Reza},
  booktitle={NAACL},
  year={2022}
}

@inproceedings{olshausen1996emergence,
  title={Emergence of simple-cell receptive field properties by learning a sparse code for natural images},
  author={Olshausen, Bruno A and Field, David J},
  booktitle={Nature},
  year={1996},
}

@inproceedings{makhzani2014k,
  title={K-sparse autoencoders},
  author={Makhzani, Alireza and Frey, Brendan},
  booktitle={ICLR},
  year={2014}
}

@inproceedings{ranzato2007sparse,
  title={Sparse feature learning for deep belief networks},
  author={Ranzato, Marc'Aurelio and Boureau, Y-Lan and Cun, Yann and others},
  booktitle={NeurIPS},
  year={2007}
}

@inproceedings{ranzato2006efficient,
  title={Efficient learning of sparse representations with an energy-based model},
  author={Ranzato, Marc'Aurelio and Poultney, Christopher and Chopra, Sumit and Cun, Yann},
  booktitle={NeurIPS},
  year={2006}
}

@inproceedings{cunningham2023sparse,
  title={Sparse autoencoders find highly interpretable features in language models},
  author={Cunningham, Hoagy and Ewart, Aidan and Riggs, Logan and Huben, Robert and Sharkey, Lee},
  booktitle={arXiv},
  year={2023}
}

@inproceedings{bricken2023towards,
  title={Towards monosemanticity: Decomposing language models with dictionary learning},
  author={Trenton Bricken and Adly Templeton and Joshua Batson and Brian Chen and Adam Jermyn and Tom Conerly and Nick Turner and Cem Anil and Carson Denison and Amanda Askell and others},
  booktitle={Transformer Circuits Thread},
  year={2023}
}

@inproceedings{elhage2022toy,
  title={Toy models of superposition},
  author={Elhage, Nelson and Hume, Tristan and Olsson, Catherine and Schiefer, Nicholas and Henighan, Tom and Kravec, Shauna and Hatfield-Dodds, Zac and Lasenby, Robert and Drain, Dawn and Chen, Carol and others},
  booktitle={arXiv},
  year={2022}
}

@inproceedings{klindt2025superposition,
  title={From superposition to sparse codes: interpretable representations in neural networks},
  author={Klindt, David and O'Neill, Charles and Reizinger, Patrik and Maurer, Harald and Miolane, Nina},
  booktitle={arXiv},
  year={2025}
}

@inproceedings{hindupur2025projecting,
  title={Projecting assumptions: The duality between sparse autoencoders and concept geometry},
  author={Hindupur, Sai Sumedh R and Lubana, Ekdeep Singh and Fel, Thomas and Ba, Demba},
  booktitle={arXiv},
  year={2025}
}

@inproceedings{costa2025flat,
  title={From Flat to Hierarchical: Extracting Sparse Representations with Matching Pursuit},
  author={Costa, Val{\'e}rie and Fel, Thomas and Lubana, Ekdeep Singh and Tolooshams, Bahareh and Ba, Demba},
  booktitle={arXiv},
  year={2025}
}

@inproceedings{rajamanoharan2024jumping,
  title={Jumping ahead: Improving reconstruction fidelity with jumprelu sparse autoencoders},
  author={Rajamanoharan, Senthooran and Lieberum, Tom and Sonnerat, Nicolas and Conmy, Arthur and Varma, Vikrant and Kram{\'a}r, J{\'a}nos and Nanda, Neel},
  booktitle={arXiv},
  year={2024}
}

@inproceedings{bussmann2024batchtopk,
  title={BatchTopK Sparse Autoencoders},
  author={Bart Bussmann and Patrick Leask and Neel Nanda},
  booktitle={NeurIPS Workshop on Scientific Methods for Understanding Deep Learning},
  year={2024}
}

@inproceedings{gao2025scaling,
  title={Scaling and evaluating sparse autoencoders},
  author={Leo Gao and Tom Dupre la Tour and Henk Tillman and Gabriel Goh and Rajan Troll and Alec Radford and Ilya Sutskever and Jan Leike and Jeffrey Wu},
  booktitle={ICLR},
  year={2025},
}

@inproceedings{nabeshima2024matryoshka,
  title={Matryoshka Sparse Autoencoders},
  author={Nabeshima, Noa},
  booktitle={LessWrong AI Alignment Forum},
  year={2024}
}

@inproceedings{engels2025not,
  title={Not All Language Model Features Are One-Dimensionally Linear},
  author={Joshua Engels and Eric J Michaud and Isaac Liao and Wes Gurnee and Max Tegmark},
  booktitle={ICLR},
  year={2025}
}

@inproceedings{engels2025decomposing,
  title={Decomposing The Dark Matter of Sparse Autoencoders},
  author={Joshua Engels and Logan Riggs Smith and Max Tegmark},
  booktitle={TMLR},
  year={2025},
}

@inproceedings{muchane2025incorporating,
  title={Incorporating Hierarchical Semantics in Sparse Autoencoder Architectures},
  author={Muchane, Mark and Richardson, Sean and Park, Kiho and Veitch, Victor},
  booktitle={arXiv},
  year={2025}
}

@inproceedings{fel2025archetypal,
  title={Archetypal {SAE}: Adaptive and Stable Dictionary Learning for Concept Extraction in Large Vision Models},
  author={Thomas Fel and Ekdeep Singh Lubana and Jacob S. Prince and Matthew Kowal and Victor Boutin and Isabel Papadimitriou and Binxu Wang and Martin Wattenberg and Demba E. Ba and Talia Konkle},
  booktitle={FICML},
  year={2025}
}

@inproceedings{hoyle2021automated,
  title={Is automated topic model evaluation broken? the incoherence of coherence},
  author={Hoyle, Alexander and Goel, Pranav and Hian-Cheong, Andrew and Peskov, Denis and Boyd-Graber, Jordan and Resnik, Philip},
  booktitle={NeurIPS},
  year={2021}
}

@inproceedings{park2024the,
  title={The Linear Representation Hypothesis and the Geometry of Large Language Models},
  author={Kiho Park and Yo Joong Choe and Victor Veitch},
  booktitle={ICML},
  year={2024}
}

@inproceedings{burkhardt2019decoupling,
  title={Decoupling sparsity and smoothness in the dirichlet variational autoencoder topic model},
  author={Burkhardt, Sophie and Kramer, Stefan},
  booktitle={JMLR},
  year={2019}
}

@inproceedings{raffel2020exploring,
  title={Exploring the limits of transfer learning with a unified text-to-text transformer},
  author={Raffel, Colin and Shazeer, Noam and Roberts, Adam and Lee, Katherine and Narang, Sharan and Matena, Michael and Zhou, Yanqi and Li, Wei and Liu, Peter J},
  booktitle={JMLR},
  year={2020}
}

@inproceedings{awasthy2025granite,
  title={Granite Embedding R2 Models},
  author={Awasthy, Parul and Trivedi, Aashka and Li, Yulong and Doshi, Meet and Bhat, Riyaz and Kumar, Vishwajeet and Yang, Yushu and Iyer, Bhavani and Daniels, Abraham and Murthy, Rudra and others},
  booktitle={arXiv},
  year={2025}
}

@inproceedings{sharma2018conceptual,
  title={Conceptual captions: A cleaned, hypernymed, image alt-text dataset for automatic image captioning},
  author={Sharma, Piyush and Ding, Nan and Goodman, Sebastian and Soricut, Radu},
  booktitle={ACL},
  year={2018}
}

@inproceedings{changpinyo2021conceptual,
  title={Conceptual 12m: Pushing web-scale image-text pre-training to recognize long-tail visual concepts},
  author={Changpinyo, Soravit and Sharma, Piyush and Ding, Nan and Soricut, Radu},
  booktitle={CVPR},
  year={2021}
}

@inproceedings{thomee2016yfcc100m,
  title={Yfcc100m: The new data in multimedia research},
  author={Thomee, Bart and Shamma, David A and Friedland, Gerald and Elizalde, Benjamin and Ni, Karl and Poland, Douglas and Borth, Damian and Li, Li-Jia},
  booktitle={Communications of the ACM},
  year={2016}
}

@inproceedings{wang2025internvl3,
  title={Internvl3. 5: Advancing open-source multimodal models in versatility, reasoning, and efficiency},
  author={Wang, Weiyun and Gao, Zhangwei and Gu, Lixin and Pu, Hengjun and Cui, Long and Wei, Xingguang and Liu, Zhaoyang and Jing, Linglin and Ye, Shenglong and Shao, Jie and others},
  booktitle={arXiv},
  year={2025}
}

@inproceedings{stammbach2023revisiting,
  title={Revisiting Automated Topic Model Evaluation with Large Language Models},
  author={Stammbach, Dominik and Zouhar, Vil{\'e}m and Hoyle, Alexander and Sachan, Mrinmaya and Ash, Elliott},
  booktitle={EMNLP},
  year={2023}
}

@inproceedings{pennington2014glove,
  title={Glove: Global vectors for word representation},
  author={Pennington, Jeffrey and Socher, Richard and Manning, Christopher D},
  booktitle={EMNLP},
  year={2014}
}

@inproceedings{mikolov2013efficient,
  title={Efficient estimation of word representations in vector space},
  author={Mikolov, Tomas and Chen, Kai and Corrado, Greg and Dean, Jeffrey},
  booktitle={arXiv},
  year={2013}
}

@inproceedings{rao2024discover,
  title={Discover-then-name: Task-agnostic concept bottlenecks via automated concept discovery},
  author={Rao, Sukrut and Mahajan, Sweta and B{\"o}hle, Moritz and Schiele, Bernt},
  booktitle={ECCV},
  year={2024},
}

@inproceedings{hermann2015teaching,
  title={Teaching machines to read and comprehend},
  author={Hermann, Karl Moritz and Kocisky, Tomas and Grefenstette, Edward and Espeholt, Lasse and Kay, Will and Suleyman, Mustafa and Blunsom, Phil},
  booktitle={NeurIPS},
  year={2015}
}

@inproceedings{cheng2010you,
  title={You are where you tweet: a content-based approach to geo-locating twitter users},
  author={Cheng, Zhiyuan and Caverlee, James and Lee, Kyumin},
  booktitle={CIKM},
  year={2010}
}

@inproceedings{maas2011learning,
  title={Learning word vectors for sentiment analysis},
  author={Maas, Andrew and Daly, Raymond E and Pham, Peter T and Huang, Dan and Ng, Andrew Y and Potts, Christopher},
  booktitle={ACL-HLT},
  year={2011}
}

@inproceedings{zhang2015character,
  title={Character-level convolutional networks for text classification},
  author={Zhang, Xiang and Zhao, Junbo and LeCun, Yann},
  booktitle={NeurIPS},
  year={2015}
}

@book{mitchell1997machine,
  author    = {Mitchell, Tom M.},
  title     = {Machine Learning},
  year      = {1997},
  publisher = {McGraw-Hill},
}

@inproceedings{rahimi2024contextualized,
  title={Contextualized Topic Coherence Metrics},
  author={Rahimi, Hamed and Mimno, David and Vigly, Jacob Hoover and Naacke, Hubert and Constantin, Camelia and Amann, Bernd},
  booktitle={EACL Findings},
  year={2024}
}

@inproceedings{harrando2021apples,
  title={Apples to apples: A systematic evaluation of topic models},
  author={Harrando, Ismail and Lisena, Pasquale and Troncy, Raphael},
  booktitle={RANLP},
  year={2021}
}

@inproceedings{chang2009reading,
  title={Reading tea leaves: How humans interpret topic models},
  author={Chang, Jonathan and Gerrish, Sean and Wang, Chong and Boyd-Graber, Jordan and Blei, David},
  booktitle={NeurIPS},
  year={2009}
}

@inproceedings{lau2014machine,
  title={Machine reading tea leaves: Automatically evaluating topic coherence and topic model quality},
  author={Lau, Jey Han and Newman, David and Baldwin, Timothy},
  booktitle={EACL},
  year={2014}
}

@inproceedings{bird2006nltk,
  title={NLTK: the natural language toolkit},
  author={Bird, Steven},
  booktitle={COLING/ACL},
  year={2006}
}

@inproceedings{wu2023effective,
  title={Effective neural topic modeling with embedding clustering regularization},
  author={Wu, Xiaobao and Dong, Xinshuai and Nguyen, Thong Thanh and Luu, Anh Tuan},
  booktitle={ICML},
  year={2023}
}

@inproceedings{deng2009imagenet,
  title={Imagenet: A large-scale hierarchical image database},
  author={Deng, Jia and Dong, Wei and Socher, Richard and Li, Li-Jia and Li, Kai and Fei-Fei, Li},
  booktitle={CVPR},
  year={2009}
}

@inproceedings{abdin2024phi,
  title={Phi-4 technical report},
  author={Abdin, Marah and Aneja, Jyoti and Behl, Harkirat and Bubeck, S{\'e}bastien and Eldan, Ronen and Gunasekar, Suriya and Harrison, Michael and Hewett, Russell J and Javaheripi, Mojan and Kauffmann, Piero and others},
  booktitle={arXiv},
  year={2024}
}

@inproceedings{zheng2025model,
  title={Model Directions, Not Words: Mechanistic Topic Models Using Sparse Autoencoders},
  author={Zheng, Carolina and Beltran-Velez, Nicolas and Karlekar, Sweta and Shi, Claudia and Nazaret, Achille and Mallik, Asif and Feder, Amir and Blei, David M},
  booktitle={arXiv},
  year={2025}
}

@inproceedings{wu2024survey,
  title={A survey on neural topic models: methods, applications, and challenges},
  author={Wu, Xiaobao and Nguyen, Thong and Luu, Anh Tuan},
  booktitle={Artificial Intelligence Review},
  year={2024}
}

@inproceedings{bowman2016generating,
  title={Generating sentences from a continuous space},
  author={Bowman, Samuel and Vilnis, Luke and Vinyals, Oriol and Dai, Andrew and Jozefowicz, Rafal and Bengio, Samy},
  booktitle={SIGNLL},
  year={2016}
}

@inproceedings{subramanian2018towards,
  title={Towards text generation with adversarially learned neural outlines},
  author={Subramanian, Sandeep and Mudumba, Sai Rajeswar and Sordoni, Alessandro and Trischler, Adam and Courville, Aaron and Pal, Chris},
  booktitle={NeurIPS},
  year={2018}
}

@inproceedings{peng2025use,
  title={Use Sparse Autoencoders to Discover Unknown Concepts, Not to Act on Known Concepts},
  author={Peng, Kenny and Movva, Rajiv and Kleinberg, Jon and Pierson, Emma and Garg, Nikhil},
  booktitle={arXiv},
  year={2025}
}

@inproceedings{angelov2020top2vec,
  title={Top2vec: Distributed representations of topics},
  author={Angelov, Dimo},
  booktitle={arXiv},
  year={2020}
}

@inproceedings{wu2021discovering,
  title={Discovering topics in long-tailed corpora with causal intervention},
  author={Wu, Xiaobao and Li, Chunping and Miao, Yishu},
  booktitle={ACL-IJCNLP (Findings)},
  year={2021}
}

@inproceedings{wu2022mitigating,
  title={Mitigating Data Sparsity for Short Text Topic Modeling by Topic-Semantic Contrastive Learning},
  author={Wu, Xiaobao and Luu, Anh Tuan and Dong, Xinshuai},
  booktitle={EMNLP},
  year={2022}
}

@inproceedings{wu2024towards,
  title={Towards the TopMost: A Topic Modeling System Toolkit},
  author={Wu, Xiaobao and Pan, Fengjun and Tuan, Luu Anh},
  booktitle={ACL},
  year={2024}
}

@inproceedings{newman2010automatic,
  title={Automatic evaluation of topic coherence},
  author={Newman, David and Lau, Jey Han and Grieser, Karl and Baldwin, Timothy},
  booktitle={NAACL-HLT},
  year={2010}
}

@inproceedings{krizhevsky2009learning,
  title={Learning multiple layers of features from tiny images},
  author={Krizhevsky, Alex and Hinton, Geoffrey and others},
  booktitle={Tech Report},
  year={2009},
}

@inproceedings{bossard14,
  title = {Food-101 -- Mining Discriminative Components with Random Forests},
  author = {Bossard, Lukas and Guillaumin, Matthieu and Van Gool, Luc},
  booktitle = {ECCV},
  year = {2014}
}

@inproceedings{xiao2010sun,
  title={Sun database: Large-scale scene recognition from abbey to zoo},
  author={Xiao, Jianxiong and Hays, James and Ehinger, Krista A and Oliva, Aude and Torralba, Antonio},
  booktitle={CVPR},
  year={2010}
}

@inproceedings{schuhmann2021laion,
  title={Laion-400m: Open dataset of clip-filtered 400 million image-text pairs},
  author={Schuhmann, Christoph and Vencu, Richard and Beaumont, Romain and Kaczmarczyk, Robert and Mullis, Clayton and Katta, Aarush and Coombes, Theo and Jitsev, Jenia and Komatsuzaki, Aran},
  booktitle={arXiv},
  year={2021}
}

@inproceedings{ilharco2021openclip,
  author={Ilharco, Gabriel and Wortsman, Mitchell and Wightman, Ross and Gordon, Cade and Carlini, Nicholas and Taori, Rohan and Dave, Achal and Shankar, Vaishaal and Namkoong, Hongseok and Miller, John and Hajishirzi, Hannaneh and Farhadi, Ali an Schmidt, Ludwig},
  title={OpenCLIP},
  booktitle={GitHub},
  year={2021},
}

@inproceedings{udandarao2024no,
  title={No" zero-shot" without exponential data: Pretraining concept frequency determines multimodal model performance},
  author={Udandarao, Vishaal and Prabhu, Ameya and Ghosh, Adhiraj and Sharma, Yash and Torr, Philip and Bibi, Adel and Albanie, Samuel and Bethge, Matthias},
  booktitle={NeurIPS},
  year={2024}
}

@inproceedings{garcia2023uncurated,
  title={Uncurated image-text datasets: Shedding light on demographic bias},
  author={Garcia, Noa and Hirota, Yusuke and Wu, Yankun and Nakashima, Yuta},
  booktitle={CVPR},
  year={2023}
}

@inproceedings{holtzman2020curious,
  title={The Curious Case of Neural Text Degeneration},
  author={Ari Holtzman and Jan Buys and Li Du and Maxwell Forbes and Yejin Choi},
  booktitle={ICLR},
  year={2020}
}

@inproceedings{zhang2024recognize,
  title={Recognize anything: A strong image tagging model},
  author={Zhang, Youcai and Huang, Xinyu and Ma, Jinyu and Li, Zhaoyang and Luo, Zhaochuan and Xie, Yanchun and Qin, Yuzhuo and Luo, Tong and Li, Yaqian and Liu, Shilong and others},
  booktitle={CVPR},
  year={2024}
}

@inproceedings{huang2023open,
  title={Open-set image tagging with multi-grained text supervision},
  author={Huang, Xinyu and Huang, Yi-Jie and Zhang, Youcai and Tian, Weiwei and Feng, Rui and Zhang, Yuejie and Xie, Yanchun and Li, Yaqian and Zhang, Lei},
  booktitle={arXiv},
  year={2023}
}

@inproceedings{khan2021stylistic,
  title={Stylistic multi-task analysis of ukiyo-e woodblock prints},
  author={Khan, Selina and van Noord, Nanne},
  booktitle={BMVC},
  year={2021}
}

@inproceedings{xiao2025flair,
  title={Flair: Vlm with fine-grained language-informed image representations},
  author={Xiao, Rui and Kim, Sanghwan and Georgescu, Mariana-Iuliana and Akata, Zeynep and Alaniz, Stephan},
  booktitle={CVPR},
  year={2025}
}

@inproceedings{kar2024brave,
  title={Brave: Broadening the visual encoding of vision-language models},
  author={Kar, O{\u{g}}uzhan Fatih and Tonioni, Alessio and Poklukar, Petra and Kulshrestha, Achin and Zamir, Amir and Tombari, Federico},
  booktitle={ECCV},
  year={2024}
}

@inproceedings{li2025unbiased,
  title={Unbiased Region-Language Alignment for Open-Vocabulary Dense Prediction},
  author={Li, Yunheng and Li, Yuxuan and Zeng, Quan-Sheng and Wang, Wenhai and Hou, Qibin and Cheng, Ming-Ming},
  booktitle={CVPR},
  year={2025}
}

@inproceedings{wu2025axbench,
  title={AxBench: Steering {LLM}s? Even Simple Baselines Outperform Sparse Autoencoders},
  author={Zhengxuan Wu and Aryaman Arora and Atticus Geiger and Zheng Wang and Jing Huang and Dan Jurafsky and Christopher D Manning and Christopher Potts},
  booktitle={ICML},
  year={2025}
}

@inproceedings{smith2025negative,
    title={Negative Results for Sparse Autoencoders On Downstream Tasks and Deprioritising SAE Research},
    author={Lewis Smith and Sen Rajamanoharan and Arthur Conmy and Callum McDougall and Janos Kramar and Tom Lieberum and Rohin Shah and Neel Nanda},
    booktitle={DeepMind Safety Research Blog Post},
    year={2025},
    howpublished={\url{https://deepmindsafetyresearch.medium.com/negative-results-for-sparse-autoencoders-on-downstream-tasks-and-deprioritising-sae-research-6cadcfc125b9}}
}

@inproceedings{kim2025interpreting,
  title={Interpreting vision transformers via residual replacement model},
  author={Kim, Jinyeong and Kim, Junhyeok and Shim, Yumin and Kim, Joohyeok and Jung, Sunyoung and Hwang, Seong Jae},
  booktitle={arXiv},
  year={2025}
}

@inproceedings{kusner2015word,
  title={From word embeddings to document distances},
  author={Kusner, Matt and Sun, Yu and Kolkin, Nicholas and Weinberger, Kilian},
  booktitle={ICML},
  year={2015}
}

@inproceedings{wiedemer2024pretraining,
  title={Pretraining Frequency Predicts Compositional Generalization of {CLIP} on Real-World Tasks},
  author={Thadd{\"a}us Wiedemer and Yash Sharma and Ameya Prabhu and Matthias Bethge and Wieland Brendel},
  booktitle={NeurIPS Workshop on Compositional Learning: Perspectives, Methods, and Paths Forward},
  year={2024}
}

@inproceedings{jiang2025interpretable,
  title={Interpretable Embeddings with Sparse Autoencoders: A Data Analysis Toolkit},
  author={Jiang, Nicholas and Sun, Xiaoqing and Dunlap, Lisa and Smith, Lewis and Nanda, Neel},
  booktitle={Mechanistic Interpretability Workshop at NeurIPS 2025},
  year={2025}
}

@inproceedings{choi2025conceptscope,
  title={ConceptScope: Characterizing Dataset Bias via Disentangled Visual Concepts},
  author={Jinho Choi and Hyesu Lim and Steffen Schneider and Jaegul Choo},
  booktitle={NeurIPS},
  year={2025}
}

@inproceedings{das2015gaussian,
  title={Gaussian LDA for topic models with word embeddings},
  author={Das, Rajarshi and Zaheer, Manzil and Dyer, Chris},
  booktitle={ACL-IJCNLP},
  year={2015}
}

@inproceedings{abaskohi2025cemtm,
  title={CEMTM: Contextual Embedding-based Multimodal Topic Modeling},
  author={Abaskohi, Amirhossein and Li, Raymond and Li, Chuyuan and Joty, Shafiq and Carenini, Giuseppe},
  booktitle={EMNLP},
  year={2025}
}
\bibliographystyle{icml2026}

\newpage
\appendix
\onecolumn
\part*{Supplementary Material}

\section{Relating Fixed-Sparsity SAEs to the CTM}
\label{sec:supp:fixedsparsitysaederivation}

Here, we relate fixed-sparsity SAEs, such as TopK \citep{gao2025scaling} and BatchTopK \citep{bussmann2024batchtopk}, to the continuous topic model (CTM) introduced in \cref*{sec:method:continuouslda}. In contrast to the $L_1$-penalty formulation analyzed in \cref*{sec:method:l1sparsitysaederivation}, fixed-sparsity SAEs enforce a hard limit on the number of active features. We show that they arise from a deterministic support-selection approximation to MAP inference under the CTM.

\subsection*{Aggregated topic activations}
Let the topic directions form the decoder matrix ($W=[\mu_1,\dots,\mu_K]\in\mathbb{R}^{d\times K}$). In the CTM, an embedding is generated by contributions ($c_n=\lambda_n w_n$), where ($w_n\sim \mathcal{N}(\mu_{z_n},\Sigma_{z_n})$). Grouping contributions by topic yields aggregated activations
\begin{equation}
a_k := \sum_{n : z_n = k} \lambda_n,
\qquad
a = (a_1,\dots,a_K)^\top \in \mathbb{R}_{\ge 0}^K.
\end{equation}
Under concentrated directions ($\Sigma_k \to 0$), $w_n \approx \mu_{z_n}$, and hence
\begin{equation}
D \mid a \ \sim\ \mathcal{N}(W a, \sigma^2 I),
\end{equation}
recovering the standard SAE reconstruction model.

\subsection*{Prior over activations}

For topic $k$, let $N_k \sim \mathrm{Pois}(\rho_d \theta_k)$ denote the number of contributions, and let each contribution strength follow $\lambda_{k,i} \sim \mathrm{Ga}(r_k, \tau_k)$. Then
\begin{equation}
a_k = \sum_{i=1}^{N_k} \lambda_{k,i}
\end{equation}
is compound-Poisson-Gamma distributed, featuring a point mass at zero (topic inactive) and a continuous density for $a_k > 0$. For the special case $r_k = 1$, this density takes the closed form
\begin{equation}
f_{a_k}(a) = e^{-(\rho_k + \tau_k a)}
\sqrt{\frac{\rho_k \tau_k}{a}}
I_1\left(2\sqrt{\rho_k \tau_k a}\right),
\qquad a>0,
\label{eq:fk-supp}
\end{equation}
where $\rho_k = \rho_d \theta_k$ and $I_1(\cdot)$ is the modified Bessel function of the first kind.
The MAP objective for a single embedding is
\begin{equation}
\mathcal{L}(a) = \frac{1}{2\sigma^2}\lVert D - W a\rVert_2^2 + \sum_{k=1}^K \bigl[-\log p(a_k)\bigr].
\label{eq:map-fixedsparsity}
\end{equation}
For $a_k = 0$, the penalty equals $\rho_k$. For $a_k>0$, the dominant term $\tau_k a_k$ induces magnitude shrinkage.

\subsection*{Deterministic support as Inference Approximation}

The exact MAP objective in \cref{eq:map-fixedsparsity} is computationally intractable, as it requires a combinatorial search over the sparse support (the ``spike'' vs.\ ``slab'') of the compound-Poisson-Gamma prior. Fixed-sparsity SAEs therefore use the encoder to find an \emph{approximate} MAP solution via deterministic support selection. Instead of sampling $N_k$ or penalizing $\lvert a_k\rvert$, the encoder chooses a subset of indices $\mathcal{S} \subset \{1,\dots,K\}$ of fixed size $\lvert\mathcal{S}\rvert = k \ll K$, sets $a_j = 0$ for $j\notin\mathcal{S}$, and computes the remaining $a_j$ directly via the encoder. This corresponds to:
\begin{equation}
-\log p(a) \approx \text{const} \quad \text{subject to } |\{k : a_k>0\}| = k,
\end{equation}
This approximation effectively replaces the complex CPG prior with a constant $L_0$ penalty. This penalty is simply the fixed hyperparameter $k$ (the number of active features), which is not learned via the objective but set externally, mirroring standard SAE training. This is, in other words, a hard constraint on the number of active topics rather than a soft prior over magnitudes.

\subsection*{Summary}

Fixed-sparsity SAEs arise from the CTM under the following simplifications:
\begin{compactitem}
\item \emph{Concentrated topic directions:} $\Sigma_k \to 0$, yielding the linear decoder $W=[\mu_1,\dots,\mu_K]$.
\item \emph{Deterministic support selection:} Approximate the intractable MAP inference under the CPG prior by directly choosing a fixed active set.
\item \emph{Low effective activity:} Use $k \ll K$, mirroring the small-$\rho_d$ regime where only a few topics contribute.
\end{compactitem}
Thus, fixed-sparsity SAEs correspond to MAP inference in the continuous topic model with a deterministic sparse-support constraint. This connection clarifies how decoder weights and sparsity levels correspond to the generative quantities in \cref*{sec:method:continuouslda}, and explains why fixed-sparsity SAEs behave as topic models in practice.

\section{SAE Pretraining}
\label{sec:supp:saepretraining}

\mypara{Text SAE}
We pretrain a foundational SAE for text on a large dataset combining sections from Wikipedia and C4 \citep{raffel2020exploring}. Concretely, we sample 240 million sections each from Wikipedia and C4 and embed them using \texttt{granite-r2} \citep{awasthy2025granite}. A section is a sequence of $n\in [1, \ldots, 10]$ consecutive sentences within a single document. In Wikipedia, we treat individual paragraphs in articles as documents and only consider paragraphs with at least 5 sentences. In C4, we consider all documents. Using the combined 480 million embedded sections, we train a BatchTopK SAE \citep{bussmann2024batchtopk} with an expansion factor of 64 (dictionary size is $49,152$ features) and 32 active features per embedding for 800,000 steps with a batch size of 4,096. The SAE achieves reconstruction $R^2 = 0.785$. 

\mypara{Image SAE}
We pretrain a SAE for images on SigLIP embeddings of 360 million images in LAION-400M \citep{schuhmann2021laion}. Concretely, we train a BatchTopK SAE for 50,000 steps with batch size 20,000 on image embeddings of VIT-B-16-SIGLIP from OpenCLIP \citep{ilharco2021openclip}. The SAE has expansion factor of 32 and 32 active features per embedding. It achieved reconstruction $R^2 = 0.872$.

\section{Detailed Description of the Topic Merging Procedure}
\label{sec:supp:topicmerging}

The goal of this procedure is merge the large number $K$ of sparse, fine-grained SAE features (\enquote{topic atoms}) into a smaller set of $K' \ll K$ distinct topics. The process uses word embeddings to cluster features that activate on semantically similar words, weighted by the frequency of feature activation. Pseudocode for the algorithm is in \cref{alg:saetopicmerging}.

\subsection{Feature Representation (Topic Atom Embeddings)}
Each SAE feature $k$ is characterized by its word emission probabilities $B_k \in \mathbb{R}^V$, representing the probability distribution $P(w|k)$ over the vocabulary. To compute a dense vector representation $\mathbf{T}_k$ for feature $k$:
\begin{enumerate}
    \item \textbf{Denoising (Top-$p$ Filtering):} We filter the distribution $B_k$ to retain only the most significant words. We select the smallest set of words whose cumulative probability mass sums to at least $p$ (in our experiments, we choose $p=0.9$).
    \item \textbf{Renormalization:} The surviving entries are renormalized to sum to 1, forming a sparse vector $\hat{B}_k$.
    \item \textbf{Projection:} We compute the weighted sum of word embeddings: $\mathbf{T}_k = \sum_{w_i \in \mathcal{V}} \hat{B}_{k,i} \cdot \mathbf{w}_i$, where $\mathbf{w}_i$ is the embedding vector for word $w_i$.
\end{enumerate}

\subsection{Weighted Clustering}
We cluster the topic atom embeddings $\{\mathbf{T}_k\}_{k=1}^K$ into $K'$ groups.
\begin{itemize}
    \item \textbf{Algorithm:} We employ $k$-means clustering on the feature embeddings.
    \item \textbf{Weighting:} The clustering is weighted by the global importance of each feature, given by $P(k)$ (the average activation value of feature $k$ across the dataset). This ensures that frequent features drive the formation of cluster centroids more than rare features.
\end{itemize}

\subsection{Topic Aggregation}
Once cluster assignments $c_k \in \{1, \dots, K'\}$ are obtained, we aggregate the features in each cluster $k'$ to form the final topic distribution $P_{k'}(w)$. This is computed as the weighted average of the original word emission distributions $B_k$, weighted by their activation frequency $P(k)$, as shown in Equation (1) of the main text.

\begin{algorithm}[H]
\SetAlgoLined
\DontPrintSemicolon
\caption{SAE Topic Merging}
\label{alg:saetopicmerging}

\KwIn{
    Word Emission Matrix $B \in \mathbb{R}^{K \times V}$ (SAE Features $\times$ Vocab), 
    Feature Frequencies $\mathbf{p}_{feat} \in \mathbb{R}^K$ (where $\mathbf{p}_{feat}[k] = P(k)$), 
    Word Embeddings $\mathbf{W} \in \mathbb{R}^{V \times D}$,
    Doc-Feature Matrix $\Theta_{csr}$,
    Cumulative Probability Threshold $p$ (e.g., 0.9), 
    Target Topic Count $K'$
}
\KwOut{Cluster Assignments $\mathcal{C}$, Aggregated Topic Distributions $\mathbf{B}'$}

\SetKwFunction{FMain}{Main}
\SetKwFunction{FSparsify}{SparsifyAndRenormalize}
\SetKwProg{Fn}{Function}{:}{}

\Fn{\FSparsify{$M$, $p$}}{
    \tcp{Keeps minimal top words summing to probability mass $p$}
    Initialize $M_{out} \leftarrow \mathbf{0}^{K \times V}$\;
    \For{each feature $k$ in $M$}{
        Sort indices $idx$ such that $M[k, idx]$ is descending\;
        Calculate cumulative sum $S = \text{cumsum}(M[k, idx])$\;
        Find cut-off index $n = \min \{i \mid S[i] > p\}$\;
        \tcp{Keep top $n$ elements, mask others}
        Set active indices $A \leftarrow idx[0 \dots n]$\;
        $M_{out}[k, A] \leftarrow M[k, A]$\;
        \tcp{Renormalize to sum to 1}
        $M_{out}[k, :] \leftarrow M_{out}[k, :] / \sum(M_{out}[k, :])$\;
    }
    \KwRet $M_{out}$\;
}

\bigskip

\SetKwProg{Pro}{Procedure}{:}{}
\Pro{\FMain}{
    \tcp{1. Filter Dead Features}
    Find active indices $\mathcal{K}_{valid} = \{k \mid \mathbf{p}_{feat}[k] > 0\}$\;
    $B_{valid} \leftarrow B[\mathcal{K}_{valid}, :]$, $P_{valid} \leftarrow \mathbf{p}_{feat}[\mathcal{K}_{valid}]$\;
    
    \tcp{2. Generate Topic Atom Embeddings}
    $B_{sparse} \leftarrow$ \FSparsify{$B_{valid}$, $p$}\;
    \tcc{Calculate weighted average of word embeddings for each SAE feature}
    $\mathbf{T} \leftarrow B_{sparse} \times \mathbf{W}$ \tcp*{$\mathbf{T}_k = \sum B_{k,i} \cdot \mathbf{w}_i$}
    
    \tcp{3. Weighted Clustering}
    Initialize KMeans with $K'$ clusters\;
    \tcc{Fit embeddings $\mathbf{T}$ weighted by feature frequency $P(k)$}
    $\mathcal{L} \leftarrow \text{KMeans}(X=\mathbf{T}, \text{weights}=P_{valid})$\;
    
    \tcp{4. Merge Topics (Aggregation)}
    Initialize $\mathcal{C}$ mapping Cluster ID $k' \to$ Feature Indices $k$\;
    Initialize $\mathbf{B}' \in \mathbb{R}^{K' \times V}$\;
    \For{$k' \leftarrow 0$ \KwTo $K'-1$}{
        Get features in cluster: $\mathcal{F}_{k'} = \{k \in \mathcal{K}_{valid} \mid \mathcal{L}[k] = k'\}$\;
        \tcc{Compute aggregated distribution: Eq. (1)}
        $Numerator \leftarrow \sum_{k \in \mathcal{F}_{k'}} (B_{valid}[k] \cdot P_{valid}[k])$\;
        $Denominator \leftarrow \sum_{k \in \mathcal{F}_{k'}} P_{valid}[k]$\;
        $\mathbf{B}'[k'] \leftarrow Numerator / Denominator$\;
    }
    
    \KwRet $\mathcal{C}, \mathbf{B}'$
}

\end{algorithm}

\section{Topic Diversity: Ratio of Unique Words}
\label{sec:supp:topicdiversitymetric}

In this section, we show the topic diversity for text-only datasets, corresponding to \cref{tab:results:textonly}, but use the ratio of unique words \citep{dieng2020topic} as the metric. The results are in \cref{tab:results:textonly:uniquewordratio}. We observe that the overall trends are similar, but the ratio of unique words decreases as the number of topics increases. This occurs because the total number of words considered for topic evaluation scales with $n \times k$, where $n = 20$ in our case (see \cref{sec:experiments}). Since we use only the top 5000 most common lemmas in each dataset for topic model training, the number of unique words is capped. Therefore, the decreasing trend in the metric is mathematically inevitable. In contrast, the WMD metric used in the main paper does not suffer from this issue and permits comparison across different numbers of topics.

\begin{table}[htpb]
\centering
\begin{tabular}{lccccc}
\toprule
& \multicolumn{5}{c}{Number of Topics} \\
 & 50 & 100 & 200 & 300 & 500 \\
\midrule
AVITM & 0.623 & 0.443 & 0.260 & 0.209 & 0.135 \\
CombinedTM & 0.637 & 0.486 & 0.385 & 0.362 & 0.233 \\
DecTM & \underline{0.784} & 0.620 & 0.428 & 0.394 & 0.284 \\
DVAE & 0.485 & 0.319 & 0.199 & 0.141 & 0.096 \\
ETM & 0.769 & \underline{0.646} & 0.509 & 0.426 & \underline{0.328} \\
FASTopic & 0.639 & 0.586 & \underline{0.527} & \underline{0.451} & 0.326 \\
NSTM & 0.406 & 0.284 & 0.207 & 0.160 & 0.057 \\
TSCTM & \textbf{0.942} & \textbf{0.806} & \textbf{0.652} & \textbf{0.536} & \textbf{0.388} \\
\midrule
SAE-TM (ours) & 0.731 & 0.592 & 0.447 & 0.369 & 0.279 \\
\bottomrule
\end{tabular}
\caption{Topic diversity scores on text-only datasets (macro-averages of all 5 datasets), using the ratio of unique words \citep{dieng2020topic}. Best values are in bold, and second-best values are underlined. Scores are not comparable for different numbers of topics, which is the case for the WMD metric in the main paper.}
\label{tab:results:textonly:uniquewordratio}
\end{table}

\section{Ablating the Background Prior Weight $\pi$}
\label{sec:supp:backgroundweightablation}

Here, we ablate the background prior weight $\pi$ introduced in \cref{sec:saetopic} (\cref{eq:saeinterpretation}) to learn more faithful SAE feature interpretations. We sweep values $\pi \in \{0.1, 0.2, \ldots, 0.9\}$ and evaluate the resulting interpretations using $C_R$ (topic relevance), $C_I$ (intruder detection), and $D$ (topic diversity) on all text datasets. The following table reports the results:

\begin{table}[h]
\centering
\resizebox{\linewidth}{!}{
\begin{tabular}{lrrrrrrrrrrrrrrr}
\toprule
$\pi$ & \multicolumn{3}{c}{50} & \multicolumn{3}{c}{100} & \multicolumn{3}{c}{200} & \multicolumn{3}{c}{300} & \multicolumn{3}{c}{500} \\
\cmidrule(l){2-4} \cmidrule(lr){5-7} \cmidrule(lr){8-10} \cmidrule(lr){11-13} \cmidrule(r){14-16}
& $C_I \uparrow$ & $C_R \uparrow$ & $D \uparrow$ & $C_I \uparrow$ & $C_R \uparrow$ & $D \uparrow$ & $C_I \uparrow$ & $C_R \uparrow$ & $D \uparrow$ & $C_I \uparrow$ & $C_R \uparrow$ & $D \uparrow$ & $C_I \uparrow$ & $C_R \uparrow$ & $D \uparrow$ \\
\midrule
0.1 & 52.69 & \underline{77.75} & 3.58 & 48.44 & 77.05 & 3.56 & 45.45 & \underline{76.67} & 3.52 & 43.19 & \underline{75.05} & 3.50 & 39.29 & \underline{72.19} & 3.47 \\
0.2 & \textbf{56.30} & \textbf{78.16} & 3.65 & 51.31 & \underline{77.70} & 3.62 & \underline{47.13} & \textbf{76.92} & 3.59 & \textbf{45.23} & \textbf{75.86} & 3.57 & \textbf{41.79} & \textbf{73.13} & 3.55 \\
0.3 & 54.53 & 77.30 & 3.67 & \textbf{52.01} & \textbf{78.25} & 3.64 & 46.16 & 75.94 & 3.60 & 43.75 & 74.28 & 3.58 & 40.56 & 71.43 & 3.57 \\
0.4 & \underline{55.30} & 77.53 & 3.70 & 50.71 & 77.03 & 3.67 & 46.77 & 75.20 & 3.64 & 43.88 & 73.41 & 3.62 & 40.95 & 71.01 & 3.60 \\
0.5 & 54.13 & 75.30 & 3.71 & \underline{51.54} & 76.43 & 3.71 & \textbf{47.34} & 74.49 & 3.68 & 44.46 & 72.44 & 3.66 & \underline{41.24} & 70.07 & 3.64 \\
0.6 & 53.26 & 74.57 & 3.73 & 50.49 & 75.87 & 3.73 & 47.11 & 73.47 & 3.71 & \underline{44.86} & 72.05 & 3.70 & 41.21 & 68.81 & 3.68 \\
0.7 & 54.38 & 75.49 & 3.78 & 50.72 & 74.65 & 3.76 & 46.63 & 72.02 & 3.75 & 43.76 & 70.19 & 3.74 & 40.55 & 67.23 & 3.72 \\
0.8 & 53.81 & 72.57 & \underline{3.80} & 49.38 & 71.22 & \underline{3.81} & 45.59 & 68.97 & \underline{3.79} & 42.99 & 67.03 & \underline{3.78} & 39.79 & 64.65 & \underline{3.77} \\
0.9 & 52.98 & 69.88 & \textbf{3.86} & 49.92 & 69.50 & \textbf{3.86} & 43.85 & 66.03 & \textbf{3.85} & 40.91 & 63.06 & \textbf{3.85} & 38.05 & 60.29 & \textbf{3.83} \\
\bottomrule
\end{tabular}
}
\caption{Ablation of the background-prior weight $\pi$ in \cref{sec:saetopic}, on text-only datasets (macro-averages across the five datasets of \cref{sec:experiments:language}). Best values per column are in bold, second-best values are underlined.}
\label{tab:supp:piablation}
\end{table}

We find that scores are stable across $\pi\in[0.2, 0.5]$, with coherence peaking near $\pi=0.2$ and diversity increasing monotonically with $\pi$. The default $\pi=0.3$ used in our main experiments lies within this stable range.

\section{Results on Topic Relevance}
\label{sec:supp:topicrelevance}

To demonstrate that the learned topics are not only coherent and diverse but also meaningful to the documents they represent, we evaluate topic relevance using an LLM (\texttt{Phi-4}). For a given document, we sample one active topic and one inactive topic and prompt the LLM to decide whether each topic is relevant to the document. If a model's assigned topics are poorly aligned with the text, the LLM will not be able to distinguish between active and inactive topics. 

\cref{tab:relevance_overall} reports the overall topic relevance accuracy averaged across all text datasets. SAE-TM outperforms all baselines across all topic numbers, and achieves a mean accuracy of 66.5\% compared to 61.7\% for the best baseline, AVITM. 

While baseline accuracy on inactive (irrelevant) topics is near 100\%, as it is clear when topics are unrelated, active but relevant topics are often broad and thus harder to correctly attribute to the document. \cref{tab:relevance_active} shows the detection accuracy exclusively on active topics. Here, the performance margin is particularly favorable for SAE-TM, which correctly identifies active topics as relevant 38.3\% of the time on average, substantially outperforming the strongest baseline (AVITM at 29.0\%).

\begin{table}[ht]
\centering
\begin{tabular}{lcccccc}
\toprule
\textbf{Model} & \textbf{50} & \textbf{100} & \textbf{200} & \textbf{300} & \textbf{500} & \textbf{Mean} \\
\midrule
AVITM      & 61.34 & 61.67 & 63.29 & 63.94 & 59.99 & 61.66 \\
CombinedTM & 58.05 & 58.61 & 58.19 & 53.37 & 55.56 & 56.73 \\
DecTM      & 57.45 & 59.21 & 59.16 & 51.95 & 52.60 & 56.08 \\
DVAE       & 49.84 & 49.88 & 49.68 & 49.79 & 49.90 & 49.74 \\
ETM        & 52.89 & 53.55 & 52.83 & 52.65 & 52.61 & 52.88 \\
Fastopic   & 55.25 & 54.62 & 55.44 & 57.14 & 57.65 & 56.26 \\
NSTM       & 50.01 & 50.78 & 51.57 & 50.97 & 52.62 & 51.36 \\
TSCTM      & 60.32 & 60.97 & 61.98 & 62.63 & 62.00 & 61.57 \\
\midrule
SAE-TM     & \textbf{64.14} & \textbf{63.74} & \textbf{66.49} & \textbf{67.54} & \textbf{69.36} & \textbf{66.51} \\
\bottomrule
\end{tabular}
\caption{Overall topic relevance accuracy (mean across text datasets). Best results are bolded.}
\label{tab:relevance_overall}
\end{table}
\begin{table}[ht]
\centering
\begin{tabular}{lcccccc}
\toprule
\textbf{Model} & \textbf{50} & \textbf{100} & \textbf{200} & \textbf{300} & \textbf{500} & \textbf{Mean} \\
\midrule
AVITM      & 27.76 & 27.87 & 31.77 & 32.66 & 25.05 & 29.00 \\
CombinedTM & 17.30 & 18.45 & 17.18 & 7.75 & 12.45 & 14.68 \\
DecTM      & 16.10 & 20.39 & 21.34 & 6.41 & 6.41 & 13.99 \\
DVAE       & 0.33 & 1.54 & 0.37 & 0.24 & 0.75 & 0.58 \\
ETM        & 8.18 & 9.27 & 7.65 & 6.95 & 6.89 & 7.66 \\
Fastopic   & 13.94 & 12.03 & 13.30 & 16.14 & 17.71 & 15.16 \\
NSTM       & 6.94 & 5.58 & 7.46 & 6.37 & 7.96 & 6.93 \\
TSCTM      & 21.42 & 22.90 & 25.16 & 26.44 & 24.88 & 24.15 \\
\midrule
SAE-TM     & \textbf{33.65} & \textbf{33.09} & \textbf{37.95} & \textbf{40.32} & \textbf{43.69} & \textbf{38.26} \\
\bottomrule
\end{tabular}
\caption{Overall detection accuracy on active topics only (mean across text datasets). Best results are bolded.}
\label{tab:relevance_active}
\end{table}

\section{Prompts}

\begin{promptbox}[Image Captioning Prompt]
You are an expert image analyst and descriptive writer specializing in creating "dense captions." Your task is to generate a single, continuous paragraph of highly detailed and comprehensive text that describes the provided image. Your description must be objective and based solely on visual evidence. \medskip

Follow this multi-step process for your analysis: \medskip

1. Holistic Overview:
Begin by establishing the overall scene. Describe the setting (e.g., urban street, natural landscape, indoor room), the time of day (e.g., midday, golden hour, night), the overall atmosphere or mood (e.g., bustling, serene, melancholic), and the general color palette. \smallskip

2. Primary Subjects and Actions:
Identify and describe the primary subject(s) in detail. If they are people, describe their apparent age, gender, clothing, posture, expression, and any actions they are performing. If they are objects or animals, describe their type, condition, color, and position. Describe the interactions between primary subjects. \smallskip

3. Secondary Elements and Background:
Detail the secondary subjects, significant objects, and the immediate background. Describe architectural elements, furniture, vehicles, flora, and fauna that populate the scene but are not the central focus. Describe their spatial relationship to the primary subjects. \smallskip

4. Fine-Grained Details and Textures:
Scrutinize the image for fine-grained details. Mention specific textures (e.g., the rough bark of a tree, the smooth surface of a metal table, the fabric weave of a coat), small, easily missed objects, text or symbols visible on signs or clothing, reflections in windows or water, and the quality of light and shadow (e.g., sharp, defined shadows indicating harsh light, or soft, diffuse light). \smallskip

5. Synthesis and Composition:
Conclude by synthesizing all observations. Briefly describe the photographic composition, such as the framing, perspective, and depth of field (e.g., a shallow depth of field blurring the background, a wide-angle shot capturing a vast landscape). \medskip

Formatting and Style Constraints: \medskip

- Output Format: Your entire output must be one single, continuous paragraph. \\
- No Line Breaks: Do not use any line breaks, newlines, or paragraph breaks (\textbackslash n). \\
- Style: Write in a descriptive, objective, and formal tone. \\
- Exclusions: \\
\hspace*{0.6em} - Do not start with phrases like "This is an image of," "The picture shows," or any similar introductory statement. \\
\hspace*{0.6em} - Do not include personal opinions, judgments, or interpretations that are not directly supported by visual evidence. \\
\hspace*{0.6em} - Do not use bullet points, lists, or headers in your final output. Your entire response must be the caption itself.\medskip

Example of Desired Output: \medskip

A vibrant and crowded marketplace unfolds under the bright, hazy sun of midday, characterized by a dominant palette of warm ochres, deep reds, and earthen browns. The central focus is a male vendor in his late fifties, wearing a light blue djellaba and a straw hat, who is carefully arranging a pyramid of colorful spices on a rough wooden stall; his face is weathered and creased in concentration. In front of him, a tourist with a backpack slung over one shoulder, clad in a khaki shirt, points at a specific spice mound while a young child clings to her hand, looking with wide eyes at a nearby stall selling intricately woven leather bags. The background is a dense tapestry of activity, with other shoppers and vendors creating a soft-focus blur of movement, set against the backdrop of ancient, reddish-pink plaster walls and arched doorways. Fine details abound, from the coarse texture of the burlap sacks holding grains and the gleam of polished brass lanterns hanging from a wooden beam, to the subtle shadows cast by the woven canopy overhead, dappling the ground in a shifting pattern of light. The composition is tight and layered, creating a deep sense of immersion and chaotic energy, capturing the scene from a slightly low, eye-level perspective that places the viewer directly within the bustling alleyway. \medskip

Your Task: \medskip

Now, analyze the following image and generate the dense caption, strictly adhering to all instructions above.
\end{promptbox}

\begin{promptbox}[Image Captioning Prompt (for Ukiyo-e Images)]
You are given an image of a Japanese woodblock print. Provide a single continuous paragraph of detailed analysis that follows these instructions: Describe the visual scene in objective, scientific, and precise language. Identify and name all visible figures, landmarks, buildings, or natural features if they are recognizable and culturally significant, and state their role in the composition. Describe the arrangement and interaction of figures, objects, and background elements. Note any inscriptions, seals, or cartouches, including their placement, without attempting speculative translation. Mention stylistic or technical aspects that are clearly visible, such as color layering, printing techniques, or use of pattern. Interpret the cultural or historical significance of the depicted subject only when directly inferable from visible attributes, without speculation or reference to things outside the image. If an element's identity or meaning is uncertain, explicitly state that it cannot be determined. Do not use subjective adjectives like "beautiful" or "ethereal", and do not mention arbitrary concepts not present in the image. Ensure that the analysis is precise, objective, culturally informed, and presented as one continuous paragraph without lists, headings, or bullet points.
\end{promptbox}

\begin{promptbox}[LLM-as-a-judge for Intruder Detection]
From the following list of words, identify the single word that does not belong with the others. The words are: \{words\}. \medskip

Your response must be only the single intruder word and nothing else.
\end{promptbox}

\begin{promptbox}[LLM-as-a-judge for Coherence Rating]
You are an expert in semantics and lexical relationships. Your task is to evaluate the coherence of the following list of words: '\{words\}'. \medskip

Coherence is how well the words belong to a single, clear, and specific category.\smallskip

\hspace*{1em} - A score of 100 means the words are extremely coherent (e.g., all are types of citrus fruits). \\
\hspace*{1em} - A score around 50 means the words are moderately coherent (e.g., all are 'vehicles' but mix cars, boats, and planes). \\
\hspace*{1em} - A score of 0 means the words are completely unrelated. \medskip

Provide your analysis as a JSON object with two keys: \enquote{rationale} and \enquote{score}.\smallskip

\hspace*{1em} - \enquote{rationale}: A brief, one-sentence explanation for your score. \\
\hspace*{1em} - \enquote{score}: An integer between 0 and 100. \medskip

Your response MUST be only the JSON object and nothing else."
\end{promptbox}

\section{Full Dataset Results}
We report the full per-dataset results, i.e.\ an expanded version of \cref{tab:results:textonly}, for text datasets in \cref{tab:supp:fullresults:text}. Additionally, we report the full per-dataset results (expanded version of \cref{tab:results:imagedataset}) for image dataset in \cref{tab:supp:fullresults:images}.

\begin{table}
\resizebox{\linewidth}{!}{
\begin{tabular}{lccccccccccccccc}
\toprule
Num.\ Topics & \multicolumn{3}{c}{50} & \multicolumn{3}{c}{100} & \multicolumn{3}{c}{200} & \multicolumn{3}{c}{300} & \multicolumn{3}{c}{500} \\
\cmidrule(l){2-4} \cmidrule(lr){5-7} \cmidrule(lr){8-10} \cmidrule(lr){11-13} \cmidrule(r){14-16}
 & $C_I \uparrow$ & $C_R \uparrow$ & $D \uparrow$ & $C_I \uparrow$ & $C_R \uparrow$ & $D \uparrow$ & $C_I \uparrow$ & $C_R \uparrow$ & $D \uparrow$ & $C_I \uparrow$ & $C_R \uparrow$ & $D \uparrow$ & $C_I \uparrow$ & $C_R \uparrow$ & $D \uparrow$ \\
\midrule
\multicolumn{16}{c}{\textbf{News-20k}} \\
\midrule
AVITM \citep{srivastava2017autoencoding} & 39.76 & 66.43 & 3.47 & 36.08 & 66.15 & 3.42 & 36.60 & 64.63 & 3.50 & 34.57 & 60.03 & 3.61 & 33.67 & 60.57 & 3.47 \\
CombinedTM \citep{bianchi2021cross} & \underline{45.64} & \underline{71.46} & \textbf{3.97} & \underline{42.26} & \underline{66.74} & \textbf{3.93} & \textbf{44.05} & \underline{70.02} & \textbf{3.96} & \textbf{48.09} & \textbf{73.51} & \textbf{3.95} & \textbf{43.46} & \underline{68.54} & \textbf{3.93} \\
DecTM \citep{wu2021discovering} & 39.68 & 60.32 & \underline{3.90} & 35.86 & 59.88 & \underline{3.87} & 33.95 & 57.93 & \underline{3.81} & 32.04 & 56.36 & \underline{3.81} & 30.12 & 53.56 & \underline{3.78} \\
DVAE \citep{burkhardt2019decoupling} & 22.32 & 27.02 & 3.66 & 18.54 & 5.34 & 3.37 & 16.71 & 3.18 & 3.37 & 17.49 & 7.05 & 3.41 & 16.56 & 2.00 & 3.38 \\
ETM \citep{dieng2020topic} & 21.64 & 17.79 & 3.16 & 18.48 & 14.65 & 3.32 & 19.47 & 10.91 & 3.44 & 18.87 & 11.79 & 3.47 & 17.88 & 8.81 & 3.52 \\
FASTopic \citep{wu2024fastopic} & 34.04 & 59.76 & 3.36 & 31.66 & 56.16 & 3.48 & 31.24 & 58.76 & 3.44 & 29.52 & 55.23 & 3.43 & 28.78 & 58.29 & 3.37 \\
NSTM \citep{zhao2021neural} & 26.04 & 42.90 & 3.22 & 27.18 & 52.10 & 3.16 & 28.71 & 56.19 & 3.11 & 29.61 & 55.90 & 3.13 & 29.80 & 54.76 & 3.12 \\
TSCTM \citep{wu2022mitigating} & 37.16 & 60.02 & 3.81 & 33.38 & 55.62 & 3.81 & 26.86 & 34.60 & 3.77 & 23.59 & 23.43 & 3.72 & 20.58 & 14.80 & 3.68 \\
\midrule
SAE-TM (ours) & \textbf{49.16} & \textbf{71.91} & 3.61 & \textbf{48.06} & \textbf{75.33} & 3.62 & \underline{41.32} & \textbf{72.93} & 3.58 & \underline{39.25} & \underline{72.35} & 3.57 & \underline{36.76} & \textbf{68.74} & 3.53 \\
\midrule
\multicolumn{16}{c}{\textbf{IMDB}} \\
\midrule
AVITM \citep{srivastava2017autoencoding} & 31.84 & \underline{67.14} & 3.07 & 30.44 & \underline{68.64} & 3.07 & 28.37 & \underline{66.23} & 3.05 & \underline{28.05} & \underline{65.60} & 3.04 & 22.73 & \underline{49.27} & 2.64 \\
CombinedTM \citep{bianchi2021cross} & \underline{37.88} & 66.54 & \textbf{3.74} & 33.64 & 63.87 & \underline{3.69} & \underline{32.10} & 60.57 & \underline{3.66} & 16.56 & 4.05 & \textbf{3.65} & 16.88 & 4.66 & \underline{3.64} \\
DecTM \citep{wu2021discovering} & 37.64 & 63.09 & 3.71 & \underline{34.98} & 61.46 & \textbf{3.77} & 19.42 & 12.41 & 3.61 & 20.65 & 18.41 & 3.57 & 16.82 & 6.35 & 3.64 \\
DVAE \citep{burkhardt2019decoupling} & 16.20 & 4.56 & 3.26 & 16.72 & 2.63 & 3.30 & 17.50 & 2.66 & 3.08 & 15.38 & 5.80 & 3.11 & 16.45 & 3.57 & 3.05 \\
ETM \citep{dieng2020topic} & 20.20 & 30.65 & 3.02 & 20.00 & 23.10 & 3.13 & 19.62 & 17.34 & 3.19 & 19.45 & 16.56 & 3.24 & 19.03 & 12.85 & 3.32 \\
FASTopic \citep{wu2024fastopic} & 31.12 & 54.42 & 1.74 & 27.66 & 46.51 & 1.94 & 26.21 & 39.91 & 2.28 & 26.08 & 51.15 & 2.64 & \underline{25.54} & 46.45 & 2.80 \\
NSTM \citep{zhao2021neural} & 16.76 & 39.99 & 2.79 & 20.10 & 41.95 & 2.73 & 18.19 & 47.99 & 2.61 & 18.50 & 46.49 & 2.60 & 17.26 & 43.40 & 2.49 \\
TSCTM \citep{wu2022mitigating} & 35.64 & 65.22 & \underline{3.73} & 27.44 & 48.23 & 3.68 & 23.30 & 33.55 & \textbf{3.69} & 20.63 & 17.34 & \underline{3.64} & 18.92 & 14.15 & \textbf{3.64} \\
\midrule
SAE-TM (ours) & \textbf{51.04} & \textbf{79.04} & 3.65 & \textbf{45.22} & \textbf{74.87} & 3.57 & \textbf{41.29} & \textbf{72.47} & 3.52 & \textbf{38.32} & \textbf{70.56} & 3.50 & \textbf{34.96} & \textbf{66.95} & 3.49 \\
\midrule
\multicolumn{16}{c}{\textbf{Yelp}} \\
\midrule
AVITM \citep{srivastava2017autoencoding} & 37.00 & 69.00 & 3.25 & 35.06 & \underline{74.59} & 3.20 & 33.14 & \underline{71.89} & 3.20 & \underline{32.69} & \underline{70.28} & 3.17 & 31.60 & \underline{68.48} & 3.15 \\
CombinedTM \citep{bianchi2021cross} & \textbf{49.36} & 72.71 & \underline{4.02} & \textbf{49.68} & 69.79 & \textbf{4.00} & \textbf{48.76} & 68.31 & \textbf{3.99} & 16.45 & 3.56 & \underline{3.75} & \textbf{43.25} & 66.44 & \textbf{3.89} \\
DecTM \citep{wu2021discovering} & 48.68 & 70.80 & \textbf{4.05} & 41.24 & 65.96 & \underline{3.94} & 29.09 & 49.99 & \underline{3.90} & 17.38 & 2.45 & 3.72 & 17.32 & 1.97 & \underline{3.77} \\
DVAE \citep{burkhardt2019decoupling} & 18.96 & 1.79 & 3.63 & 16.92 & 3.59 & 3.21 & 16.49 & 1.32 & 3.15 & 17.19 & 1.53 & 3.10 & 16.99 & 0.98 & 2.97 \\
ETM \citep{dieng2020topic} & 19.80 & 29.69 & 3.12 & 19.92 & 20.50 & 3.21 & 19.32 & 16.08 & 3.32 & 18.37 & 13.46 & 3.38 & 17.84 & 10.40 & 3.44 \\
FASTopic \citep{wu2024fastopic} & 30.44 & 45.88 & 3.54 & 30.06 & 42.09 & 3.44 & 30.69 & 44.41 & 3.44 & 28.48 & 38.82 & 3.30 & 27.60 & 36.58 & 3.23 \\
NSTM \citep{zhao2021neural} & 21.96 & 50.99 & 3.03 & 24.94 & 55.63 & 2.95 & 24.53 & 51.28 & 2.92 & 25.11 & 54.12 & 2.97 & 24.67 & 49.17 & 3.06 \\
TSCTM \citep{wu2022mitigating} & 48.56 & \underline{74.13} & 3.97 & 33.48 & 51.91 & 3.85 & 26.85 & 32.53 & 3.80 & 23.23 & 21.88 & \textbf{3.76} & 20.98 & 14.32 & 3.74 \\
\midrule
SAE-TM (ours) & \underline{48.80} & \textbf{81.11} & 3.58 & \underline{47.60} & \textbf{80.68} & 3.55 & \underline{43.87} & \textbf{77.33} & 3.52 & \textbf{40.56} & \textbf{75.86} & 3.51 & \underline{37.93} & \textbf{72.19} & 3.51 \\
\midrule
\multicolumn{16}{c}{\textbf{Dailymail}} \\
\midrule
AVITM \citep{srivastava2017autoencoding} & 47.24 & 82.67 & 3.48 & 41.68 & \underline{82.31} & 3.49 & 38.31 & 78.73 & 3.47 & 36.83 & \underline{75.87} & 3.49 & 33.35 & \underline{72.88} & 3.45 \\
CombinedTM \citep{bianchi2021cross} & 54.80 & 79.07 & \textbf{4.09} & \underline{50.14} & 77.83 & \textbf{4.03} & 44.51 & 74.25 & \textbf{3.99} & \underline{40.27} & 71.69 & \textbf{3.99} & \underline{38.28} & 69.46 & \textbf{3.99} \\
DecTM \citep{wu2021discovering} & 51.64 & 77.12 & \underline{4.07} & 48.58 & 81.32 & 3.97 & \underline{45.70} & \underline{84.08} & 3.86 & 38.71 & 72.61 & 3.81 & 36.07 & 67.97 & 3.76 \\
DVAE \citep{burkhardt2019decoupling} & 31.60 & 77.88 & 0.69 & 18.34 & 36.02 & 3.20 & 16.73 & 1.61 & 3.41 & 16.59 & 2.45 & 3.28 & 15.70 & 1.13 & 3.37 \\
ETM \citep{dieng2020topic} & 23.28 & 35.52 & 3.16 & 20.38 & 30.42 & 3.24 & 20.08 & 31.40 & 3.27 & 20.61 & 29.61 & 3.26 & 20.60 & 27.32 & 3.24 \\
FASTopic \citep{wu2024fastopic} & 33.64 & 64.16 & 3.02 & 31.94 & 68.03 & 2.99 & 29.70 & 64.24 & 2.42 & 28.87 & 64.14 & 2.50 & 30.32 & 63.69 & 2.47 \\
NSTM \citep{zhao2021neural} & 26.12 & 57.57 & 2.74 & 24.66 & 58.75 & 2.68 & 24.16 & 50.88 & 2.59 & 23.90 & 54.36 & 2.55 & 22.00 & 47.59 & 2.37 \\
TSCTM \citep{wu2022mitigating} & \underline{57.32} & \textbf{86.09} & 4.05 & 45.60 & 73.35 & \underline{3.98} & 35.15 & 49.24 & \underline{3.88} & 30.51 & 36.22 & \underline{3.81} & 25.26 & 24.48 & \underline{3.77} \\
\midrule
SAE-TM (ours) & \textbf{67.16} & \underline{83.64} & 3.82 & \textbf{62.06} & \textbf{83.58} & 3.79 & \textbf{60.78} & \textbf{85.43} & 3.74 & \textbf{57.51} & \textbf{85.63} & 3.70 & \textbf{54.15} & \textbf{83.58} & 3.67 \\
\midrule
\multicolumn{16}{c}{\textbf{Twitter}} \\
\midrule
AVITM \citep{srivastava2017autoencoding} & 37.76 & 60.00 & 3.53 & 35.42 & 58.95 & 3.60 & \underline{36.54} & 59.38 & 3.15 & \underline{34.76} & 56.55 & 3.06 & \underline{34.05} & 55.71 & 2.92 \\
CombinedTM \citep{bianchi2021cross} & 16.80 & 61.41 & 0.40 & 16.72 & 61.04 & 0.48 & 18.36 & 44.61 & 1.66 & 17.55 & 60.80 & 0.70 & 17.08 & 60.42 & 0.75 \\
DecTM \citep{wu2021discovering} & 16.60 & 61.15 & 0.59 & 16.48 & \underline{61.33} & 0.49 & 16.50 & \underline{61.27} & 0.55 & 17.29 & \underline{61.09} & 0.65 & 17.22 & \underline{61.10} & 0.73 \\
DVAE \citep{burkhardt2019decoupling} & 17.12 & 1.00 & \textbf{3.78} & 16.26 & 0.64 & \textbf{3.74} & 16.94 & 2.67 & 3.07 & 16.55 & 6.61 & 3.11 & 16.78 & 4.79 & 3.06 \\
ETM \citep{dieng2020topic} & 23.48 & 28.14 & 2.94 & 22.52 & 24.95 & 2.99 & 21.59 & 19.53 & 3.03 & 21.54 & 18.13 & 3.08 & 20.78 & 13.48 & 3.18 \\
FASTopic \citep{wu2024fastopic} & 14.00 & 60.97 & 0.00 & 36.16 & 49.18 & 3.74 & 31.89 & 39.52 & 3.47 & 33.36 & 40.76 & 3.45 & 15.04 & 61.01 & 0.00 \\
NSTM \citep{zhao2021neural} & 17.76 & 6.66 & 3.58 & 17.50 & 2.98 & 3.65 & 17.44 & 2.09 & \textbf{3.71} & 15.83 & 1.57 & 3.52 & 14.89 & 4.04 & 3.15 \\
TSCTM \citep{wu2022mitigating} & \underline{44.36} & \underline{63.29} & \underline{3.78} & \underline{39.16} & 56.92 & 3.65 & 35.40 & 50.09 & 3.64 & 32.89 & 38.11 & \underline{3.59} & 22.67 & 17.25 & \textbf{3.67} \\
\midrule
SAE-TM (ours) & \textbf{55.40} & \textbf{70.57} & 3.72 & \textbf{54.44} & \textbf{75.58} & \underline{3.68} & \textbf{45.88} & \textbf{70.39} & \underline{3.67} & \textbf{41.87} & \textbf{66.69} & \textbf{3.65} & \textbf{38.67} & \textbf{64.66} & \underline{3.62} \\
\bottomrule
\end{tabular}
}
\caption{Full results (expanded version of \cref{tab:results:textonly}) for text datasets.}
\label{tab:supp:fullresults:text}
\end{table}

\begin{table}
\centering
\resizebox{\linewidth}{!}{
\begin{tabular}{lccccccccccccccc}
\toprule
Num.\ Topics & \multicolumn{3}{c}{50} & \multicolumn{3}{c}{100} & \multicolumn{3}{c}{200} & \multicolumn{3}{c}{300} & \multicolumn{3}{c}{500} \\
\cmidrule(l){2-4} \cmidrule(lr){5-7} \cmidrule(lr){8-10} \cmidrule(lr){11-13} \cmidrule(r){14-16}
 & $C_I \uparrow$ & $C_R \uparrow$ & $D \uparrow$ & $C_I \uparrow$ & $C_R \uparrow$ & $D \uparrow$ & $C_I \uparrow$ & $C_R \uparrow$ & $D \uparrow$ & $C_I \uparrow$ & $C_R \uparrow$ & $D \uparrow$ & $C_I \uparrow$ & $C_R \uparrow$ & $D \uparrow$ \\
\midrule
\multicolumn{16}{c}{\textbf{CIFAR-100}} \\
\midrule
AVITM \citep{srivastava2017autoencoding} & 33.28 & 76.77 & 3.43 & 33.64 & 78.06 & 3.40 & 30.94 & 77.33 & 3.36 & 30.90 & \underline{75.96} & 3.36 & 29.86 & \underline{73.80} & 3.36 \\
CombinedTM \citep{bianchi2021cross} & \underline{45.64} & 78.18 & 3.80 & \underline{43.50} & 79.34 & 3.77 & 16.12 & 8.93 & 3.83 & 16.16 & 8.25 & 3.86 & 16.36 & 7.48 & \underline{3.87} \\
DecTM \citep{wu2021discovering} & 41.40 & 71.98 & \underline{3.96} & 40.80 & 61.64 & \textbf{4.07} & 36.76 & 57.34 & \textbf{4.03} & 16.23 & 6.20 & \textbf{3.89} & 16.69 & 5.56 & \textbf{3.92} \\
DVAE \citep{burkhardt2019decoupling} & 15.96 & 2.42 & 3.86 & 15.94 & 2.87 & 3.56 & 16.05 & 5.98 & 3.22 & 16.07 & 4.82 & 3.18 & 15.99 & 5.02 & 3.10 \\
ETM \citep{dieng2020topic} & 20.28 & 37.21 & 3.44 & 19.74 & 29.15 & 3.52 & 19.00 & 23.14 & 3.62 & 18.57 & 20.20 & 3.65 & 17.93 & 14.21 & 3.71 \\
FASTopic \citep{wu2024fastopic} & 38.36 & 75.50 & 3.52 & 32.62 & 75.10 & 3.40 & 32.13 & 71.67 & 3.49 & 33.15 & 73.13 & 3.45 & \underline{32.60} & 71.95 & 3.49 \\
NSTM \citep{zhao2021neural} & 20.36 & 66.31 & 2.79 & 15.40 & 62.92 & 2.65 & 17.90 & 68.50 & 2.74 & 17.66 & 67.24 & 2.61 & 17.70 & 70.10 & 2.47 \\
TSCTM \citep{wu2022mitigating} & \textbf{46.04} & \textbf{83.50} & \textbf{3.98} & \textbf{43.88} & \textbf{83.23} & \underline{3.89} & \textbf{40.94} & \underline{80.32} & \underline{3.87} & \underline{36.37} & 68.51 & \underline{3.86} & 28.57 & 44.79 & 3.86 \\
\midrule
SAE-TM (ours) & 43.16 & \underline{82.52} & 3.51 & 40.18 & \underline{83.05} & 3.50 & \underline{38.99} & \textbf{83.15} & 3.49 & \textbf{38.56} & \textbf{82.25} & 3.48 & \textbf{37.65} & \textbf{82.02} & 3.50 \\
\midrule
\multicolumn{16}{c}{\textbf{FOOD-101}} \\
\midrule
AVITM \citep{srivastava2017autoencoding} & 27.16 & \underline{77.73} & 3.47 & 25.14 & \underline{75.30} & 3.37 & 24.13 & \underline{73.47} & 3.33 & 23.60 & \underline{72.53} & 3.32 & 23.10 & \underline{71.09} & 3.35 \\
CombinedTM \citep{bianchi2021cross} & \textbf{35.68} & 76.15 & 3.77 & \underline{32.06} & 72.56 & 3.74 & 16.08 & 30.87 & 3.61 & 16.45 & 27.71 & 3.63 & \textbf{31.08} & 68.38 & \underline{3.75} \\
DecTM \citep{wu2021discovering} & 27.48 & 62.72 & \textbf{3.86} & 26.04 & 56.41 & \textbf{3.85} & 26.00 & 57.55 & \textbf{3.83} & 16.14 & 22.99 & \underline{3.64} & 15.41 & 16.34 & 3.71 \\
DVAE \citep{burkhardt2019decoupling} & 17.84 & 2.52 & 3.66 & 15.86 & 6.58 & 3.61 & 17.11 & 13.44 & 3.12 & 16.38 & 15.12 & 3.11 & 16.84 & 10.61 & 2.92 \\
ETM \citep{dieng2020topic} & 20.52 & 50.49 & 3.45 & 18.96 & 41.18 & 3.52 & 19.38 & 33.55 & 3.56 & 18.58 & 30.08 & 3.60 & 18.12 & 24.96 & 3.65 \\
FASTopic \citep{wu2024fastopic} & 32.04 & 64.32 & 3.55 & 30.86 & 64.93 & 3.58 & \underline{29.34} & 62.74 & 3.63 & \underline{28.13} & 60.34 & 3.62 & 28.42 & 59.77 & 3.61 \\
NSTM \citep{zhao2021neural} & 17.32 & 64.66 & 2.64 & 16.76 & 63.80 & 2.54 & 18.43 & 64.57 & 2.43 & 16.55 & 64.04 & 2.43 & 17.52 & 65.78 & 2.70 \\
TSCTM \citep{wu2022mitigating} & 31.92 & 72.46 & \underline{3.82} & 30.94 & 74.10 & \underline{3.75} & 27.43 & 58.92 & \underline{3.76} & \underline{24.05} & 44.02 & \textbf{3.77} & 21.30 & 35.00 & \textbf{3.76} \\
\midrule
SAE-TM (ours) & \underline{34.12} & \textbf{84.60} & 3.46 & \textbf{33.40} & \textbf{83.80} & 3.48 & \textbf{32.05} & \textbf{83.24} & 3.48 & \textbf{30.67} & \textbf{82.96} & 3.47 & \underline{29.75} & \textbf{81.43} & 3.46 \\
\midrule
\multicolumn{16}{c}{\textbf{SUN397}} \\
\midrule
AVITM \citep{srivastava2017autoencoding} & 34.92 & 84.03 & 3.44 & 34.46 & 83.59 & 3.43 & 31.81 & 81.60 & 3.38 & 31.43 & \underline{79.86} & 3.37 & 31.50 & \underline{78.41} & 3.39 \\
CombinedTM \citep{bianchi2021cross} & 48.92 & 79.42 & 3.95 & \textbf{48.12} & 79.84 & \underline{3.91} & \underline{44.36} & 77.52 & \underline{3.88} & 37.35 & 71.97 & \textbf{3.97} & 17.11 & 23.13 & 3.71 \\
DecTM \citep{wu2021discovering} & 36.92 & 73.08 & \underline{3.95} & 39.32 & 80.85 & 3.85 & 33.76 & 61.64 & \textbf{4.02} & 16.71 & 30.38 & 3.58 & 16.12 & 19.32 & \underline{3.72} \\
DVAE \citep{burkhardt2019decoupling} & 16.96 & 3.44 & 3.93 & 16.20 & 2.86 & 3.73 & 16.79 & 6.30 & 3.19 & 16.89 & 7.96 & 3.08 & 15.74 & 7.06 & 3.09 \\
ETM \citep{dieng2020topic} & 21.20 & 48.91 & 3.37 & 20.38 & 43.76 & 3.46 & 21.01 & 35.90 & 3.53 & 20.31 & 29.81 & 3.57 & 19.29 & 25.45 & 3.63 \\
FASTopic \citep{wu2024fastopic} & 37.36 & 72.66 & 3.52 & 39.48 & 74.00 & 3.64 & 39.56 & 73.61 & 3.69 & \underline{38.39} & 73.44 & 3.71 & \underline{38.70} & 74.01 & 3.71 \\
NSTM \citep{zhao2021neural} & 19.16 & 59.15 & 2.85 & 20.80 & 71.25 & 2.93 & 21.36 & 71.74 & 2.92 & 20.13 & 69.62 & 2.86 & 18.92 & 71.10 & 2.83 \\
TSCTM \citep{wu2022mitigating} & \underline{49.68} & \underline{85.62} & \textbf{4.02} & 46.12 & \underline{86.58} & \textbf{3.93} & 39.98 & \underline{82.85} & 3.88 & 33.69 & 62.11 & \underline{3.87} & 28.07 & 42.68 & \textbf{3.90} \\
\midrule
SAE-TM (ours) & \textbf{50.44} & \textbf{88.03} & 3.68 & \underline{46.74} & \textbf{90.16} & 3.64 & \textbf{44.74} & \textbf{90.21} & 3.65 & \textbf{43.47} & \textbf{89.95} & 3.63 & \textbf{42.24} & \textbf{89.85} & 3.62 \\
\bottomrule
\end{tabular}
}
\caption{Full results (expanded version of \cref{tab:results:imagedataset}) for image datasets.}
\label{tab:supp:fullresults:images}
\end{table}


\end{document}